\documentclass{article}
\usepackage[final]{corl_2022} %
\usepackage{graphicx}
\usepackage{comment}
\usepackage{amsmath,amssymb} %
\usepackage{color}
\usepackage{hyperref}
\usepackage{textcomp, gensymb}
\usepackage{adjustbox}
\usepackage{symbols}
\usepackage{multirow}
\usepackage{multicol}
\usepackage{fontawesome}
\usepackage{subcaption}
\usepackage{wrapfig}
\usepackage{longtable}
\usepackage{booktabs}
\usepackage{tablefootnote}
\usepackage{arydshln}

\newcommand{\jw}[1]{{\color{black}#1}}

\definecolor{gray}{rgb}{0.5, 0.5, 0.5}
\definecolor{purple}{rgb}{0.5, 0.0, 0.5}
\definecolor{amber}{rgb}{1.0, 0.75, 0.0}
\definecolor{airforceblue}{rgb}{0.36, 0.54, 0.66}
\definecolor{darkmidnightblue}{rgb}{0.0, 0.2, 0.4}
\definecolor{darkolivegreen}{rgb}{0.33, 0.42, 0.18}
\definecolor{otterbrown}{rgb}{0.4, 0.26, 0.13}
\newcommand{\thetitle}{Last-Mile Embodied Visual Navigation\xspace}

\title{\thetitle} %

\author{
  Justin Wasserman$^{1*\dagger}$, Karmesh Yadav$^{2}$, Girish Chowdhary$^{1\dagger}$,\\ \textbf{Abhinav Gupta}$^{3}$, \textbf{Unnat Jain}$^{2*}$\\
  $^{1}$University of Illinois at Urbana-Champaign,\\$^{2}$Meta AI Research, $^{3}$Carnegie Mellon University\\
 
}
\newcommand\nnfootnote[1]{%
  \begin{NoHyper}
  \renewcommand\thefootnote{}\footnote{#1}%
  \addtocounter{footnote}{-1}%
  \end{NoHyper}
}
\begin{document}
\nnfootnote{$^*$equal technical contribution; $^\dagger$corresponding authors}
\maketitle

\begin{abstract}
    Realistic long-horizon tasks like image-goal navigation involve exploratory and exploitative phases. Assigned with an image of the goal, an embodied agent must explore to \textit{discover the goal}, \ie, search efficiently using learned priors. Once the goal is discovered, the agent must accurately calibrate the \textit{last-mile of navigation} to the goal. As with any robust system, \textit{switches} between exploratory goal discovery and exploitative last-mile navigation enable better recovery from errors. 
    Following these intuitive guide rails, we propose
    \ourmethod to improve the performance of existing image-goal navigation systems.
    Entirely complementing prior methods, we focus on last-mile navigation and leverage the underlying geometric structure of the problem with neural descriptors.
    With simple but effective switches, we can easily connect \ourmethod with heuristic, reinforcement learning, and neural modular policies. On a standardized image-goal navigation benchmark~\cite{hahn2021no}, we improve performance across policies, scenes, and episode complexity, raising the state-of-the-art from 45\% to 55\% success rate. Beyond photorealistic simulation, we conduct real-robot experiments in three physical scenes and find these improvements to transfer well to real environments. 
    Code and results: 
    \url{https://jbwasse2.github.io/portfolio/SLING}
    
\end{abstract}    
\vspace{-4mm}

\section{Introduction}
\label{sec:intro}
Imagine you are at a friend's home and you want to find the couch you have seen in your friend's photo. At first, you use semantic priors \ie priors about the semantic structure of the world, to navigate to the living room (a likely place for the couch). But as soon as you get the first glimpse of the couch, you implicitly estimate the relative position of the couch, use intuitive geometry, and navigate towards it. We term the latter problem, of navigating to a visible object or region, as last-mile navigation.

The field of visual navigation has a rich history. Early approaches used hand-designed features with geometry for mapping followed by standard planning algorithms. But such an approach fails to capture the necessary semantic priors that could be learned from data. Therefore, in recent years, we have seen more efforts and significant advances in capturing these priors for semantic navigation tasks such as image-goal~\cite{ZhuARXIV2016,chaplotNeuralTopologicalSLAM2020,meng2020scaling,hahn2021no,al2022ZER,yadav2022OVRL} and object-goal navigation~\cite{chaplot2020object,batra2020objectnav,habitat-challenge-2020,ye2021auxiliary}. The core idea is to train a navigation policy using reinforcement or imitation learning and capture semantics. But in an effort to capture the semantic priors, these approaches almost entirely bypass the underlying geometric structure of the problem, specifically when the object or view of interest has already been discovered.

One can argue that last-mile navigation can indeed be learned from data itself.  We agree that, in principle, it can be. However, we argue and demonstrate that an unstructured local policy for last-mile navigation is either (a) sample inefficient (billions of frames in an RL framework~\cite{wijmans2019dd}) or (b) biased and generalize poorly when learned from offline demonstrations (due to distributional shift~\cite{RossAISTATS2011,levine2020offline}). Therefore, our solution is to revisit the basics! We propose \textbf{S}witchable \textbf{L}ast-Mile \textbf{I}mage-Goal \textbf{N}avi\textbf{g}ation (\ourmethod) -- a simple yet very effective geometric navigation
system
and associated switches. Our approach can be combined with any off-the-shelf learned policy that uses semantic priors to explore the scene. As soon as the object or view of interest is detected, the \ourmethod switches to the geometric navigation
system.
We observe that \ourmethod provides significant performance gains across baselines, simulation datasets, episode difficulty, and real-world scenes.

Our key contributions are:
(1) A general-purpose \exploit~\jw{ system} and switches, that we connect with five diverse \explore methods, leading to improvements across the board.
(2) A new state-of the-art of 54.8\% success \ie a huge jump of 21.8\% \vs published work~\cite{al2022ZER} and 9.2\% \vs a concurrent preprint~\cite{yadav2022OVRL}, on the most widely-tested fold (Gibson-curved) of the AI Habitat image-goal navigation benchmark~\cite{hahn2021no};
(3) Extensive robot experiments of image-goal navigation in challenging settings with improved performance over a neural, modular policy~\cite{hahn2021no} trained on real-world data~\cite{zhou2018stereo}.

\begin{figure*}[t]
    \centering
    {
        \phantomsubcaption\label{fig:task_overview}
        \phantomsubcaption\label{fig:policy_overview}
    }
    \setlength{\tabcolsep}{20pt}
    \begin{tabular}{cc}
        \includegraphics[height=0.4\textwidth]{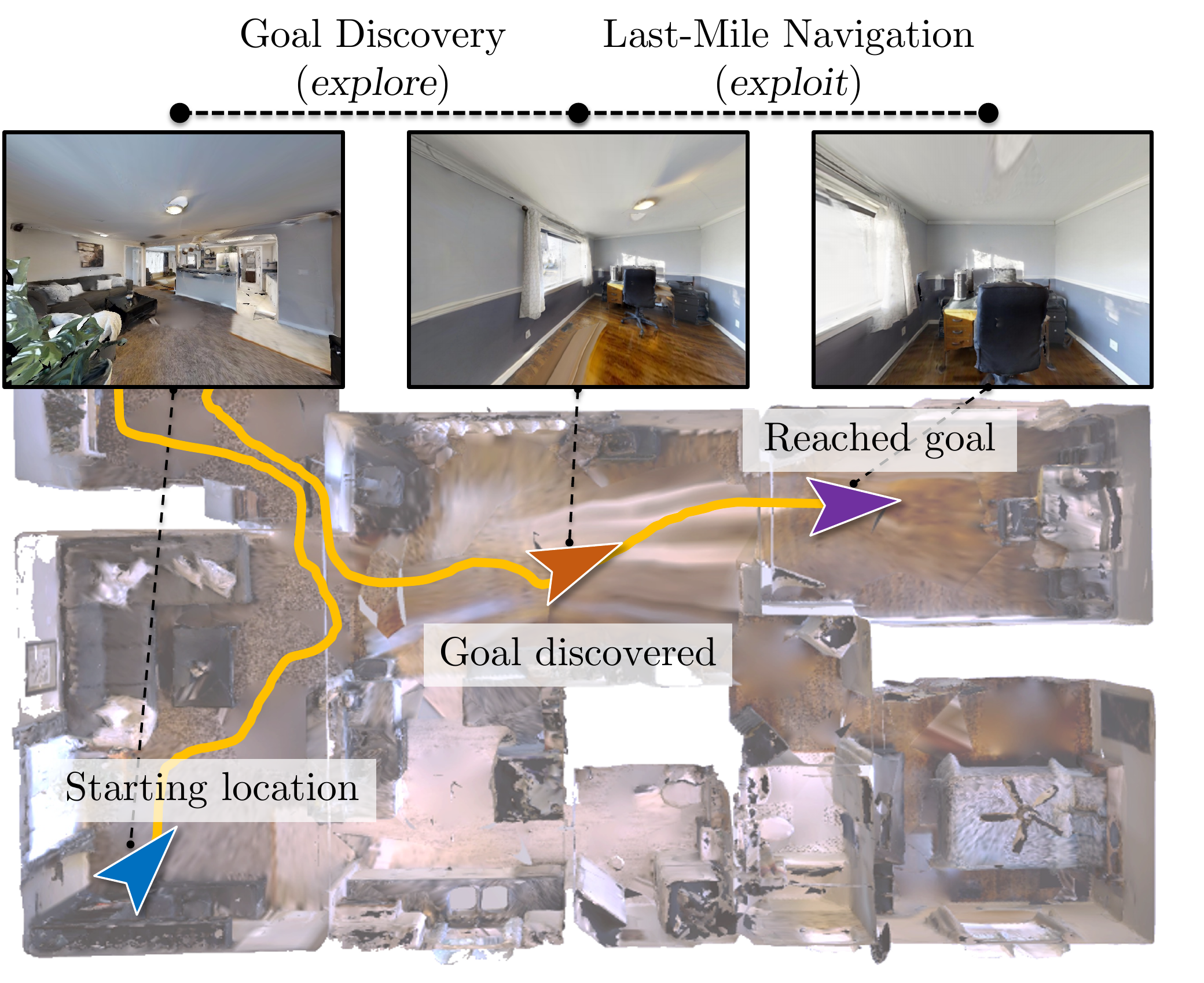} &
        \includegraphics[height=0.4\textwidth]{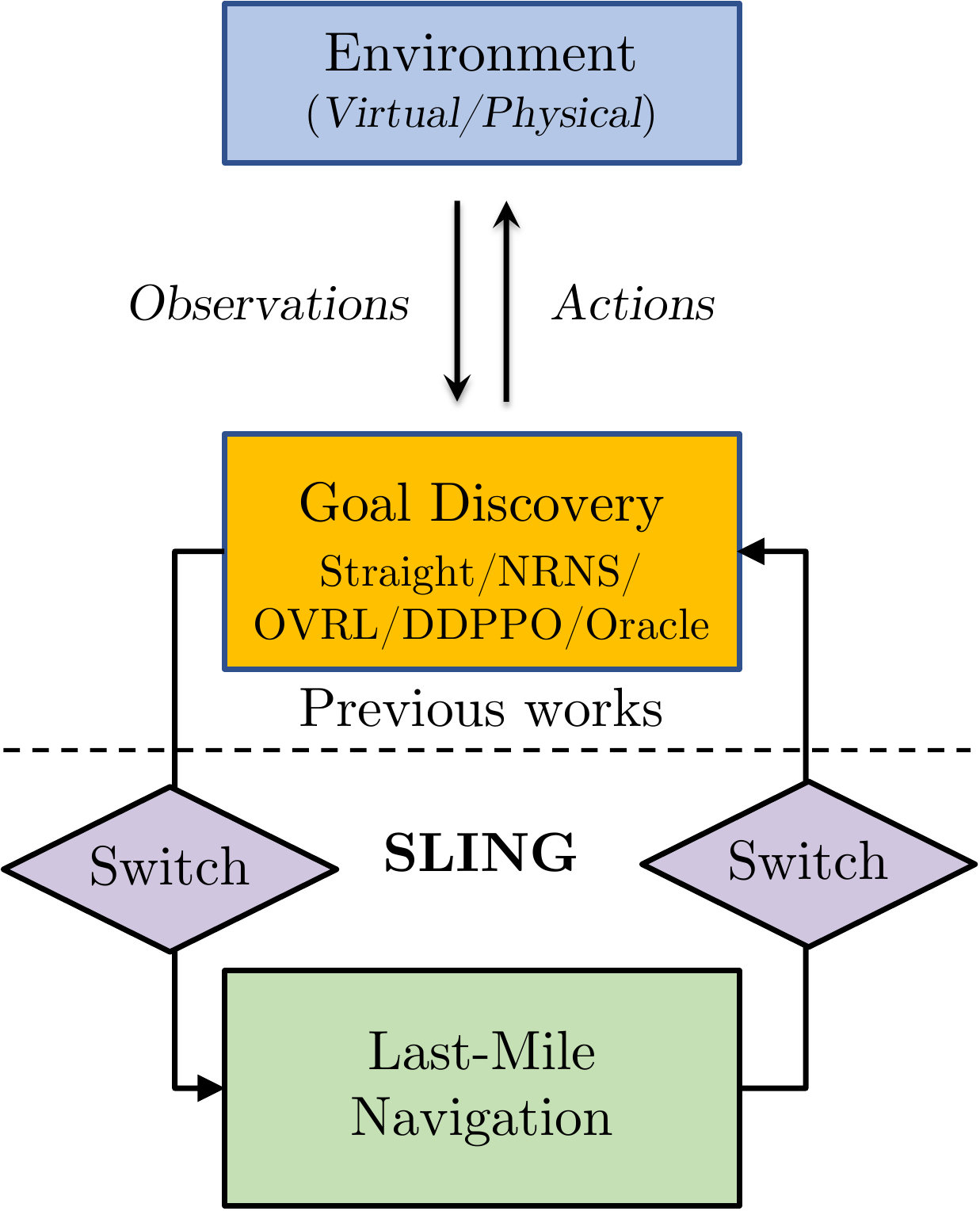}\\
        (a) Phases in Image-Goal Navigation & (b) Policy Overview
    \end{tabular}
    \vspace{2mm}
    \caption{\textbf{Switchable Last-Mile Image-Goal Navigation}. (a) Long-horizon semantic tasks such as image-goal navigation involves exploratory discovery of goals and exploitative last-mile navigation, (b) An overview of \ourmethod that allows for \textit{switching} between policies from prior work and our \exploit~system.
    }
    \label{fig:teaser}
    \vspace{-6mm}
\end{figure*}
\vspace{-2mm}
\section{Related Work}
\vspace{-2mm}
\label{sec:related}
Prior work in visual navigation and geometric 3D vision is pertinent to \ourmethod.

\noindent\textbf{Embodied navigation.}
Anderson~\etal~\cite{anderson2018evaluation} formalized different goal definitions and metrics for the evaluation of embodied agents. In point-goal navigation, relative coordinates of the goal are available (either at all steps~\cite{habitat19iccv,wijmans2019dd,habitat-challenge-2019,ye2020auxiliary,WeihsJain2020Bridging} or just at the start of an episode~\cite{habitat-challenge-2020,habitat-challenge-2021,datta2020integrating}).
Successful navigation to a point-goal could be done without semantic scene understanding, as seen by competitive depth-only agents~\cite{habitat19iccv,wijmans2019dd}.
Semantic navigation entails identifying the goal through an image (image-goal~\cite{hahn2021no,ZhuARXIV2016,AllenAct}), acoustic cues (audio-goal~\cite{chen2019audio,gan2021threedworld}), or a category label (object-goal~\cite{batra2020objectnav,habitat-challenge-2020}). Several extensions of navigation include language-conditioned navigation following~\cite{anderson2018vision,Suhr2019CerealBar,hahn2020you,raychaudhuri2021language}, social navigation~\cite{srivastava2022behavior,teach,puig2021watchandhelp,igibson,darpino2021socialnav}, and multi-agent tasks~\cite{jain2019CVPRTBONE,JainWeihs2020CordialSync,thomason2020vision,jain2021gridtopix,liu2021iros,liu2021cooperative}. However, each of these build-off single-agent navigation and benefit from associated advancements. For more embodied tasks and paradigms, we refer the reader to a recent survey~\cite{deitke2022retrospectives}.
In this work, we focus on image-goal navigation in visually rich environments.

\noindent\textbf{Image-goal navigation.}
Chaplot~\etal~\cite{chaplotNeuralTopologicalSLAM2020} introduced a modular and hierarchical method for navigating to an image-goal that utilizes a topological map memory. 
Kwon~\etal~\cite{kwon2021visual} introduced a memory representation based on image similarity, which in turn is learned in an unsupervised fashion from unlabeled data and the agent's observed images.
Following up on~\cite{chaplotNeuralTopologicalSLAM2020}, NRNS~\cite{hahn2021no} improves the topological-graph-based architecture and open-sourced a public dataset and IL and RL baselines~\cite{wijmans2019dd,chaplotNeuralTopologicalSLAM2020} within AI Habitat. This dataset has been adopted for standardized evaluation~\cite{al2022ZER,yadav2022OVRL}. 
ZER~\cite{al2022ZER} focuses on transferring an image-goal navigation policy to other navigation tasks. In a concurrent preprint, Yadav~\etal~\cite{yadav2022OVRL} utilize self-supervised pretraining~\cite{caron2021emerging} to improve an end-to-end visual RL policy~\cite{wijmans2019dd} for the image-goal navigation benchmark. Our contributions are orthogonal to the above and can be easily combined with them, as we demonstrate in~\secref{sec:exp}.

Beyond simulation, SLING finds relevance to the rich literature of navigating to an image-goal on physical robots. 
Meng~\etal~\cite{meng2020scaling} utilize a neural reachability estimator and a local controller based on a Riemannian Motion Policy framework to navigate to image-goals.
Hirose~\etal~\cite{Hirose2019deepMPC} train a deep model predictive control policy to follow a trajectory given by a sequence of images while being robust to variations in the environment. Even in outdoor settings, meticulous studies have shown great promise, based on negative mining, graph pruning, and waypoint prediction~\cite{shahViNGLearningOpenWorld2020} and utilizing geographic hints for kilometer-long navigations~\cite{shah2022viking}.
Complementing this body of work, \ourmethod tackles image-goal navigation in challenging indoor settings, without needing any prior data of the test environment (similar to~\cite{hahn2021no,yadav2022OVRL,chaplotNeuralTopologicalSLAM2020}) \ie during evaluation no access to information (trajectories, GPS, or top-down maps) in the test scenes is assumed.

\noindent\textbf{Last-mile navigation.} 
The works included above focus primarily on goal discovery.
In contrast, recent works have also identified `last-mile' errors that occur when the goal is in sight of or close to the agent.
For multi-object navigation, Wani~\etal~\cite{wani2020multion,patel2021comon} observed a two-fold improvement when allowing an error budget for the final `found' or `stop' actions. Chattopadhyay~\etal~\cite{chattopadhyay2021robustnav} found the last step of navigation to be brittle 
\ie small perturbations lead to severe failures. 
Ye~\etal~\cite{ye2021auxiliary} identified last-mile errors as a prominent error mode (10\% of the failures) in object-goal navigation. 
However, none of these works address the problem with the last-mile of navigation. From a study inspired by~\cite{wani2020multion}, we infer that better (or more tolerant to error) last-mile navigation can indeed lead to better performance in the image-goal navigation task (details in Appendix~\ref{subsec:stop}).

\noindent\textbf{Connections to 3D vision.}
The objective of our last-mile navigation \jw{system} is to predict the relative camera pose between two images \ie agent's view and image-goal.
To this end, pose estimation of a calibrated camera from 3D-to-2D point correspondences connects our embodied navigation task to geometric 3D computer vision.
The Perspective-n-Point (PnP) formulation, with extensive research and efficient solvers~\cite{fischler1981random,lu2000fast,lepetit2009epnp}, fits this use case perfectly. 
To find an accurate PnP solution, locating correspondences between the local features of the two images is critical.
We utilize SuperGlue~\cite{sarlin2020superglue} which is based on correspondences learned via attention graph neural nets and partial assignments.
We defer details of PnP and finding correspondences to \secref{subsec:exploit}, to make the approach self-sufficient.
Notably, different from related
works in 3D vision~\cite{agarwala2022planeformers,Rockwell2022,elbanani2020unsupervisedrr},
we apply \jw{\ourmethod} to sequential decision-making in embodied settings, particularly, image-goal navigation. To take policies to the real world, we utilized robust SLAM methods~\cite{murORB2,labbe2019rtab} for local odometry and pose estimation, which has also been found reliable by prior works in sim-to-real~\cite{vln-pano2real,biDirTruong2021,Sim2RealHabitat}.

\vspace{-2mm}
\section{\ourmethod}
\vspace{-2mm}
\label{sec:app}
In this section, we begin with an overview of the task and the entire pipeline of \ourmethod. We then discuss the implementations for \explore, our proposed \jw{system} for \exploit, and switches to easily combine it with prior works. 
While we explain key design choices in the main paper, a supplementary description and a list of hyperparameters, for effective reproducibility, is deferred to Appendix~\ref{subsec:supp-implement}.

\subsection{Overview}
\label{subsec:overview}
We follow the image-goal navigation task benchmark by Hahn~\etal~\cite{hahn2021no} (similar to the prior formulations~\cite{ZhuARXIV2016,chaplotNeuralTopologicalSLAM2020}). The agent observes an RGB image $\agentimage$, a depth map $\agentdepth$, and the image-goal $\goalimage$.
The agent can sample actions from $\mathcal{A} =$ \{\texttt{move forward}, \texttt{turn right}, \texttt{turn left}, \texttt{stop}\}.
The \texttt{stop} action terminates the episode.

As shown in \figref{fig:task_overview}, we divide image-goal navigation into -- a \explore and a \exploit phase. 
In the \explore phase, the agent is responsible for discovering the goal \ie navigating close enough for the goal to occupy a large portion of the egocentric observation (`goal discovered' image). 
\figref{fig:policy_overview} shows how the control flows between \jw{our system}. If the \textit{explore$\rightarrow$exploit} switch isn't triggered, learning-based exploration will continue.
Otherwise, if the \textit{explore$\rightarrow$exploit} switch triggers, the agent's observations now overlap with the image-goal and the control flows to the \exploit~\jw{system}.
We find that a one-sided flow (as attempted in~\cite{hahn2021no,chaplotNeuralTopologicalSLAM2020}) from \textit{explore$\rightarrow$exploit} is too optimistic.
Therefore, we introduce symmetric switches, including one that flows control back to \explore.

\vspace{-2mm}
\subsection{\Explore}
\vspace{-2mm}
\label{subsec:explore} %
We can combine our versatile \exploit~\jw{system} and switching mechanism with any prior method. These prior methods are previously suggested solutions to image-goal navigation. We demonstrate this with five diverse \explore (GD) implementations.

\noindent\textbf{Straight~\cite{chen2018learning}.} A simple, heuristic exploration where the agent moves forward and unblocks itself, if stuck, by turning right (similar to an effective exploration baseline in~\cite{chen2018learning}).

\noindent\textbf{\nrnsgdfull~(\nrnsgd)~\cite{hahn2021no}.} Exploratory navigation is done by proposing waypoints in navigable areas (determined utilizing the agent's depth mask), history is maintained using a topological map, and processed using graph neural nets.
The minimum cost waypoint is chosen utilizing outputs from a distance prediction network. More details are given in Appendix~\ref{subsec:supp-explore} and~\cite{hahn2021no}.

\noindent\textbf{Decentralized Distributed PPO (\ddppogd)~\cite{wijmans2019dd}.} An implementation of PPO~\cite{schulman2017proximal} for photorealistic simulators where rendering is the computational bottleneck. This has been a standard end-to-end deep RL baseline in prior works, across tasks~\cite{ye2020auxiliary,hahn2021no, al2022ZER,yadav2022OVRL,khandelwal2022simple}.

\noindent\textbf{Offline Visual Representation Learning (\ovrlgd)~\cite{yadav2022OVRL}.} A DDPPO network, with its visual encoder  pretrained using self-supervised pretext tasks~\cite{caron2021emerging} on images obtained from 3D scans~\cite{eftekhar2021omnidata}. 

\noindent\textbf{Environment-State Distance Prediction (\oraclegd).} To quantify the effect of errors coming from the \explore phase, we devise an upper bound.
This is a privileged variant of \nrnsgd that 
accesses the
ground-truth distances from the environment, exclusively for the \explore phase. For fine details of its construction, particularly, how we curtail this to be an oracle explorer and not an oracle policy, see Appendix~\ref{subsec:supp-explore}.

\vspace{-2mm}
\subsection{\Exploit}
\label{subsec:LMNav}
\begin{figure}[t]
    \centering
    \vspace{-4mm}
    \includegraphics[width=\textwidth]{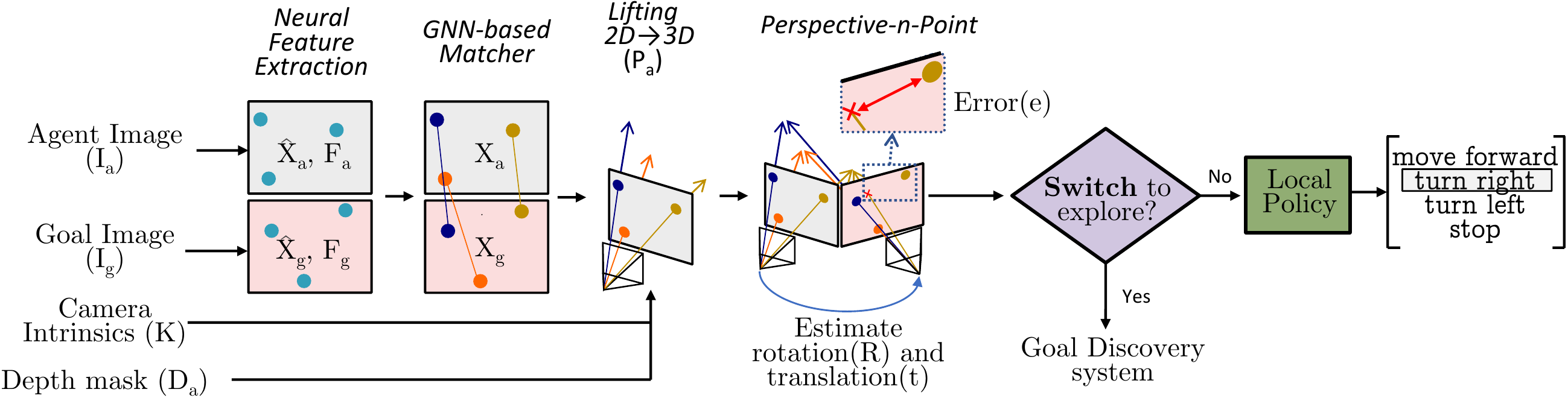}
    \caption{\textbf{\Exploit~\jw{system}.}
    Neural keypoint feature descriptors are extracted and matched to obtain correspondences between the agent's view and the image-goal. The geometric problem of estimating the relative pose between the agent and goal view is solved using efficient perspective-n-point. A \textit{exploit$\rightarrow$explore} switch, if triggered, flows control back to the \explore phase. Else, the estimations are fed into a local policy head to decide the agent's actions. 
    }
    \label{fig:lm_overview}
    \vspace{-3mm}
\end{figure}

\vspace{-2mm}
 The proposed \exploit module transforms the agent's observations and image-goal into actions that take the agent closer to the goal. The steps are shown in \figref{fig:lm_overview} and detailed next.

\noindent\textbf{\Extractor.}
We first transform the agent's RGB $\agentimage$ to local features $(\agentfeatureshat,\agentdescriptors)$, where $\agentfeatureshat \in\mathbb{R}^{n_a \times 2}$ are the positions and $\agentdescriptors \in\mathbb{R}^{n_a \times k}$ are the visual descriptors in the agent's image. Here, $n_a$ is the number of detected local features and $k$ is the length of each descriptor. Similarly, $\goalimage$ leads to features $(\goalfeatureshat,\goaldescriptors)$, where $\goalfeatureshat \in\mathbb{R}^{n_g \times 2}$ and $\goaldescriptors \in\mathbb{R}^{n_g \times k}$ with $n_g$ local features in the image-goal.
Following DeTone~\etal~\cite{detone2018superpoint}, we adopt an interest-point detector, pretrained on synthetic data followed by cross-domain homography adaptation (here, $k=256$).

\noindent\textbf{\Matcher.} From extracted features $(\agentfeatureshat,\agentdescriptors)$ and $(\goalfeatureshat,\goaldescriptors)$, we predict matched subsets $\agentfeatures\in\mathbb{R}^{n \times 2}$ and $\goalfeatures\in\mathbb{R}^{n \times 2}$. The matching is optimized to have $\agentfeatures$ and $\goalfeatures$ correspond to the same point.
We utilize an attention-based graph neural net (GNN) that tackles partial matches and occlusions well using an optimal transport formulation, following Sarlin~\etal~\cite{sarlin2020superglue}. The above neural feature extractor and GNN-based matcher help enjoy benefits of learning-based methods, particularly, those \textit{pretrained} on large offline visual data without needing online, end-to-end finetuning. The geometric components, relying on these neural features, are described next.

\label{subsec:exploit}

\noindent\textbf{Lifting Points from 2D$\rightarrow$3D.} Next, the agent's 2D local features are lifted to 3D with respect to the agent's coordinate frame \ie $\agentfeaturestd \in\mathbb{R}^{n \times 3}$. 
This is done by utilizing the camera intrinsic matrix $\camintrinsics$ (particularly, principle point $p_x, p_y$ and focal lengths $f_x, f_y$) and the corresponding depth values for each position in $\agentfeatures$, say $\depthV \in\mathbb{R}^{n}$. The $i^{\text{th}}$ row of $\agentfeaturestd$ is calculated as 
    \begin{align}
        \agentfeaturestd\left(i,:\right) &= \left(\frac{\agentfeatures(i,1) - \principleX}{\focalX} * \depthV(i), \frac{\agentfeatures(i,2) - \principleY}{\focalY} * \depthV(i), \depthV(i)\right),
    \end{align}
where $\agentfeatures(i,1)$ and $\agentfeatures(i,2)$ correspond to the $x$ and $y$ coordinate of $i^{\text{th}}$ feature in $\agentfeatures$, respectively. Formally, $\depthV(i) := \agentdepth\left(\agentfeatures(i,1), \agentfeatures(i,2)\right)$. 

\noindent\textbf{Perspective-n-Point.}
The objective of the next step \ie Perspective-n-Point (PnP) is to find the rotation and translation between the agent and goal camera pose that minimized reprojection error.
Concretely, for a given rotation matrix $\rot \in\mathbb{R}^{3 \times 3}$ and translation vector $\trans\in\mathbb{R}^{3}$, the 3D positions $\agentfeaturestd$ of local features can be reprojected from the coordinate system of the agent to that of the goal camera:
\begin{align}\label{eq:proj-and-error}
        \begin{bmatrix}
        \goalfeaturest\\
        1
        \end{bmatrix}
        =
        \camintrinsics
        \begin{bmatrix}
        \rot | \trans
        \end{bmatrix}
        \begin{bmatrix}
        \agentfeaturestd\\
        1
        \end{bmatrix};
        \hspace{8mm}
        \text{Reprojection error }e = \|\goalfeaturest - \goalfeatures\|^{2}_{2}.
\end{align}
where $\goalfeaturest$ are the reprojected positions.
Minimizing the reprojection error $e$, via ePnP~\cite{lepetit2009epnp} and RANSAC~\cite{fischler1981random} (to handle outliers), we obtain the predicted rotation and translation. The reprojection is visualized in \figref{fig:lm_overview}, where the agent's {\color{amber}amber} point is lifted and reprojected in goal camera coordinates. The reprojection is different from its correspondence in the goal image.

\noindent\textbf{Estimating Distance and Heading to Goal.}
The predicted translation $\trans$ can help calculate the distance $\dist= \|\trans\|_2$ from the agent to the goal. Similarly, the heading $\head$ from the agent to the goal can be obtained from the dot product of the unit vectors along the optical axis (of the agent's view) and $\trans$. Concretely, $\head = \text{sgn}(t[1]) * \arccos{(\trans \cdot \optic / \|\trans\|_2 \|\optic\|_2})$.
The sign comes from $t[1]$ which points along the axis perpendicular to the agent's optical axis but parallel to the ground.
The sign is particularly important when calculating the heading as it distinguishes between the agent turning right or left.
    
\noindent\textbf{Local policy.}
Finally, the distance $\dist$ and heading $\head$ between the agent's current position to the estimated goal are utilized to estimate actions in the action space $\mathcal{A}$ to reach the goal. Following accurate implementations~\cite{Chaplot2020Explore,hahn2021no}, we adopt a local metric map to allow the agent to heuristically avoid obstacles and move towards the goal. For further details, see Appendix~\ref{subsec:supp-implement}.

\vspace{-2mm}
\subsection{Switches}
\vspace{-2mm}
\label{subsec:switches}

We define simple but effective switches between the two phases of \explore (\textit{explore}) and \exploit (\textit{exploit}). The \textit{explore$\rightarrow$exploit} switch is triggered if the number of correspondences $n > n_{\text{th}}$, where $n_{\text{th}}$ is a set threshold. This indicates that the agent's image has significant overlap with the image-goal, so control can flow to the \exploit phase. 
We find that this simple switch performs better than training a specific deep net to achieve the same (variations attempted in~\cite{hahn2021no,chaplotNeuralTopologicalSLAM2020,meng2020scaling}).
For \textit{exploit$\rightarrow$explore}, if the optimization for $\rot, \trans$ (see \equref{eq:proj-and-error}) fails or if the predicted distance is greater than $d_{\text{th}}$ (tuned to 4m), the agent returns to the \explore phase.

\vspace{-2mm}
\section{Experiments}
\label{sec:exp}
\vspace{-2mm}
We report results for image-goal navigation both in photorealistic simulation and real-world scenes. 
\vspace{-4mm}
\subsection{Data and Evaluation}
\label{sec:data_and_eval}
\vspace{-2mm}
We evaluate image-goal navigation policies on the benchmark introduced by Hahn~\etal~\cite{hahn2021no} and follow their evaluation protocol and folds.
The benchmark consists of numerous folds: \{Gibson~\cite{xia2018gibson}, MP3D~\cite{chang2017matterport3d}\}$\times$\{straight, curved\}$\times$\{easy, medium, hard\}.
For a direct comparison to prior work~\cite{chaplotNeuralTopologicalSLAM2020,hahn2021no,al2022ZER,yadav2022OVRL} that reports primarily on `Gibson-curved' fold, we follow the same in the main paper. \jw{Consistent performance trends are seen in `Gibson-straight' and in the MP3D folds as well. These results are deferred to Appendix~\ref{subsec:supp-straightDataset} and Appendix~\ref{subsec:supp-curved}.}
Performance on image-goal navigation is chiefly evaluated via two metrics -- percentage of successful episodes (\success) and success weighted by inverse path length (\spl)~\cite{anderson2018evaluation}. For top-performing baselines, we also include the average distance to the goal at the end of the episode in Appendix~\ref{subsec:dtoGoal}.
The objective of the image-goal navigation task is to execute \texttt{stop} within $1m$ of the goal location. The agent is allowed $500$ steps.
\vspace{-2mm}
\subsection{Methods}
\label{subsec:baselines}
\vspace{-2mm}
We compare our last-mile navigation with several standardized baselines~\cite{bain1995framework,wijmans2019dd,hahn2021no}. Note that field-of-view, rotation amplitude, \etc vary across baselines and we adopt the respective settings for fair comparison (implementation details of \ourmethod are in Appendix~\ref{subsec:supp-implement}). Prior methods use a mix of sensors including RGB, depth, and agent pose, but no dense displacement vector to the goal. 
\jw{While we did include the most relevant baselines in~\secref{subsec:explore}, we also compare \ourmethod to several other image-goal solvers. This includes imitation learning baselines such as Behavior Cloning (BC) w/ Spatial Memory~\cite{bain1995framework} 
 and BC w/ Gated Recurrent Unit~\cite{bain1995framework,ChoARXIV2015}. We also compare to established reinforcement learning baselines -- DDPPO~\cite{wijmans2019dd} and Offline Visual Representation Learning (OVRL)~\cite{yadav2022OVRL}. OVRL also makes use of pretraining using a self-supervised objective. Finally, we compare to related modular baselines include NRNS~\cite{hahn2021no} and Zero Experience Replay (ZER)~\cite{al2022ZER}. We defer a detailed discussion of these baselines to Appendix~\ref{subsec:supp-explore}.}

\noindent\textbf{\ourmethod \& Ablations.} For a comprehensive empirical study, we combine \ourmethod with \straightgd, \nrnsgd, \ddppogd, \ovrlgd, and \oraclegd (see \secref{subsec:explore} for details). \jw{We also introduce a neural baseline, DDPPO-LMN, a DDPPO model trained to perform last-mile navigation.}

Further, we include clear ablations to show the efficacy of the components of our method and robustness to realistic pose and depth sensor noise: \\
\textbullet~\textit{w/ MLP switch:} instead of \ourmethod's explore$\rightarrow$exploit switch (that utilizes geometric structure), if a MLP\footnote{trained over an offline dataset of expert demonstrations, where adjacent nodes in a topological graph (that they maintain) are considered positives} detects similarity between the agent and goal images (as in~\cite{hahn2021no}).\\
\textbullet~\textit{w/o Recovery:} if the \textit{exploit$\rightarrow$explore} switch is removed \ie one-sided flow of control.\\
\textbullet~\textit{w/o Neural Features:} if the neural features~\cite{detone2018superpoint} are replaced with traditional features~\cite{sift}.\\
\textbullet~\textit{w/ Pose Noise:} add noise to pose that emulates real-world sensors~\cite{Chaplot2020Explore,pyrobot2019} (same as~\cite{chaplotNeuralTopologicalSLAM2020,hahn2021no}).\\
\textbullet~\textit{w/ Depth Noise:} imperfect depth by adopting the Redwood Noisy Depth model~\cite{Choi_2015_CVPR} in AI Habitat.\\
\textbullet~\textit{w/ \oraclegd:} privileged baseline where NRNS-GD can access  ground-truth distances to move the agent closer to the goal during exploration (see \secref{subsec:explore} and Appendix~\ref{subsec:supp-explore}).\\
\textbullet~\textit{w/ \oracledet:} privileged last-mile \jw{system} with perfect displacement from agent to goal \\
\textbullet~\textit{w/ \oraclept:} privileged baseline where local policy can teleport agent to the goal prediction\\

\begin{table}[t]
\centering
\caption{\textbf{Results for `Gibson-curved' episodes} Note the significant gains by adding \ourmethod to prior works. Consistent trends are seen in `Gibson-straight' (Appendix~\ref{subsec:supp-straightDataset}) and MP3D-curved episodes (Appendix~\ref{subsec:supp-curved}).
}
\resizebox{\textwidth}{!}{
\begin{tabular}{l@{\hskip 4mm} l c c c c c c c c}
    \toprule
    &&\multicolumn{2}{c}{Overall}&\multicolumn{2}{c}{Easy}&\multicolumn{2}{c}{Medium}&\multicolumn{2}{c}{Hard}\\
    & Method & Succ$\uparrow$ & SPL$\uparrow$ & Succ$\uparrow$ & SPL$\uparrow$ & Succ$\uparrow$ & SPL$\uparrow$ & Succ$\uparrow$ & SPL$\uparrow$\\
    \midrule
    1& BC w/ Spatial Memory~\cite{bain1995framework} & 1.3 & 1.1 & 3.1 & 2.5 & 0.8 & 0.7 & 0.2 & 0.1\\
    2& BC w/ GRU~\cite{bain1995framework,ChoARXIV2015} & 1.7 & 1.3 & 3.6 & 2.8 & 1.1 & 0.9 & 0.5 & 0.3\\
    3& DDPPO~\cite{wijmans2019dd} (from~\cite{hahn2021no}) & 15.7 & 12.9 & 22.2 & 16.5 & 20.7 & 18.5 & 4.2 & 3.7\\
    4& NRNS~\cite{hahn2021no} & 21.7 & 8.1 & 31.4 & 10.7 & 22.0 & 8.2 & 11.9 & 5.4\\
    5 & ZER~\cite{al2022ZER} & 33.0 & 23.6 & 48.0 & 34.2 & 36.0&  25.9 &15.1& 10.8\\
    6 & OVRL~\cite{yadav2022OVRL} & 45.6 & 28.0 & 53.6 & 31.7 & 47.6 & 30.2 & 35.6 & 21.9\\
    \jw{7} & \jw{DDPPO-LMN} + \ovrlgd & 44.3 & 30.1 & 52.4 & 36.6 & 48.6 &32.6 &31.9 &21.2\\
    8 & \ourmethod + \straightgd & 31.0 & 12.8 & 39.2 & 14.3 & 33.0 & 14.3 & 21.0 & 9.9\\
    9 & SLING + \ddppogd & 37.9 & 22.8 & 52.2 & 32.7 & 42.2 & 25.2 & 19.4 & 10.5\\
    10 & \ourmethod + \nrnsgd & 43.5 & 15.1 & 58.7 & 17.4 & 47.0 & 17.4 & 25.0 & 10.5\\
    11 & \ourmethod + \ovrlgd & \textbf{54.8} & \textbf{37.3} & \textbf{65.4} & \textbf{45.7} & \textbf{59.5} & \textbf{40.6} & \textbf{39.6} & \textbf{25.5}\\

    \bottomrule
\end{tabular}
}  
\vspace{-3mm}
\label{tab:main-curved}
\end{table}

\begin{table}[t]
\centering
\caption{\textbf{Ablations on `Gibson-curved' episodes.} Both switches are key to \ourmethod's performance. 
\ourmethod is resilient to sensor noise. \jw{Similar trends can be observed over ablations performed with \ovrlgd in Appendix~\ref{subsec:supp-curved}.} \jw{The privileged} last-mile navigation \jw{system} establishes an upper bound for last-mile navigation. Even with \oraclegd, performance improves if \ourmethod is added. }

\resizebox{\textwidth}{!}{
\begin{tabular}{l@{\hskip 4mm} l c c c c c c c c}
    \toprule
    &&\multicolumn{2}{c}{Overall}&\multicolumn{2}{c}{Easy}&\multicolumn{2}{c}{Medium}&\multicolumn{2}{c}{Hard}\\
    &Method & Succ$\uparrow$ & SPL$\uparrow$ & Succ$\uparrow$ & SPL$\uparrow$ & Succ$\uparrow$ & SPL$\uparrow$ & Succ$\uparrow$ & SPL$\uparrow$\\
    \midrule
    1& NRNS~\cite{hahn2021no} & 21.7 & 8.1 & 31.4 & 10.7 & 22.0 & 8.2 & 11.9 & 5.4\\
    \midrule
    2& \ourmethod + \nrnsgd & 43.5 & 15.1 & 58.7 & 17.4 & 47.0 & 17.4 & 25.0 & 10.5\\
    3& w/ MLP Switch & 42.5 & 14.8 & 55.4 & 16.7 & 47.3 & 17.3 & 24.9 & 10.5\\
    4& w/ MLP Switch w/o Recovery & 31.5 & 11.5 & 45.6 & 14.3 & 32.8 & 12.9 & 16.1 & 7.3\\
    5& w/o Neural Features & 33.7 & 11.3 & 47.5 & 13.5 & 35.9 & 13.0 & 17.7 & 7.5\\
    6& w/ Pose Noise & 43.7 & 14.3 & 58.6 & 16.1 & 47.6 & 16.8 & 24.9 & 10.1\\
    7& w/ Pose \& Depth Noise & 43.5&14.0 & 56.9 & 15.9 & 47.2 & 15.9 & 26.6 & 10.3\\
    \midrule
    \multicolumn{10}{c}{\textit{Privileged Last-Mile Navigation}}\\
    \midrule
    8 &  {{\color{gray}w/ \oraclept}} &{\color{gray} 45.1 }&{\color{gray}17.8 }&{\color{gray}60.8}&{\color{gray}21.2}&{\color{gray}48.7}&{\color{gray}20.3}&{\color{gray}25.8}&{\color{gray}12.1}\\
    9 & {{\color{gray}w/ \oracledet}} &{\color{gray}53.3}&{\color{gray}19.3}&{\color{gray}72.3}&{\color{gray}23.4}&{\color{gray}57.1}&{\color{gray}21.6}&{\color{gray}30.5}&{\color{gray}13.1}\\
    10 & {{\color{gray}w/ \oracledet \& \oraclept}} &{\color{gray} 53.7 }&{\color{gray} 22.4 }&{\color{gray} 72.6 }&{\color{gray} 27.7 }&{\color{gray} 57.6 }&{\color{gray} 24.7 }&{\color{gray} 31.0 }&{\color{gray}14.9}\\

    \midrule
    \multicolumn{10}{c}{\textit{Privileged Goal Discovery}}\\
    \midrule
    11& {\color{gray}NRNS + Oracle-GD \textit{(upper bound)}} & {\color{gray}67.7}& {\color{gray}60.2}& {\color{gray}68.5} & {\color{gray}58.4} & {\color{gray}71.2} & {\color{gray}63.7} & {\color{gray}63.5} & {\color{gray}58.7}\\
    12& {\color{gray}SLING + Oracle-GD \textit{(upper bound)}} & {\color{gray}86.2} & {\color{gray}74.8}& {\color{gray}85.9} & {\color{gray}72.2} & {\color{gray}88.6} & {\color{gray}77.7} & {\color{gray}84.3} & {\color{gray}74.6}\\
    \bottomrule
\end{tabular}
}
\vspace{-5mm}
\label{tab:ablation-curved}
\end{table}

\vspace{-2mm}
\subsection{Quantitative Results}
\vspace{-2mm}
\label{subsec:exp_results}
In the following, we include takeaways based on the results in \tabref{tab:main-curved} and \tabref{tab:ablation-curved}. 

\noindent\textbf{State-of-the-art performance.} As \tabref{tab:main-curved} details, \ourmethod~+~\ovrlgd  outperforms a suite of IL, RL, and neural modular baselines. The Gibson-curved fold is widely adopted in prior works and hence the focus of the main paper.
With a 54.8\% overall success and 37.3 SPL we are the best-performing method on the benchmark, improving success rate by 21.8\% \vs ZER and 9.2\% \vs OVRL (`overall success' column of rows 5, 6, \& 11). In Appendix~\ref{subsec:pan}, we also demonstrate state-of-the-art performance when panoramic images are used.

\noindent\textbf{\ourmethod works across methods.} Using switches, we add our \exploit~\jw{system} to DDPPO~\cite{wijmans2019dd}, NRNS~\cite{hahn2021no}, and OVRL~\cite{yadav2022OVRL}, and observe gains across the board.
As shown in~\tabref{tab:main-curved}, \ourmethod improves the success rate of DDPPO by 22.2\%, NRNS by 21.8\%, and OVRL by 9.2\% (rows 3 \& 9, 4 \& 10, 6 \& 11, respectively). Quite surprisingly, \ourmethod even with simple straight exploration, can outperform deep IL, RL\jw{, and modular baselines.} (rows 1, 2, 3, \jw{4}, \& 8).

\noindent\textbf{\ourmethod outperforms neural policies for \exploit.}
\ourmethod surpasses DDPPO trained over 400M steps for last-mile navigation by 10.5\% on success rate (rows 7 \& 11).

\noindent\textbf{\ourmethod succeeds across scene datasets.}
\jw{Similar improvements are also seen in MP3D scenes -- adding \ourmethod to OVRL improves success by 5.1\%. Further details and results can be found in Appendix~\ref{subsec:supp-curved}}.

\noindent\textbf{\ourmethod is resilient to sensor noise.} As shown in rows 6 \& 7 of \tabref{tab:ablation-curved}, minor drops in performance are observed despite challenging noise in pose and depth sensors -- SPL successively reduces 15.1$\rightarrow$14.3$\rightarrow$14.0\% (rows 2$\rightarrow$6$\rightarrow$7).

\noindent\textbf{Geometric switches are better.} Performance reduces if we swap out \ourmethod's explore$\rightarrow$exploit switch with the MLP switch of NRNS~\cite{hahn2021no}. The effect is exasperated when \ourmethod's exploit$\rightarrow$explore switch is also removed, leading to a drop of 12\% (\tabref{tab:ablation-curved}, rows 2 \& 4). The neural features utilized in \ourmethod are useful, as seen by comparing rows 2 and 5. Further, over a set of $6500$ image pairs, we evaluate the accuracy of switches. \ourmethod's explore$\rightarrow$exploit switch is 92.0\% accurate and MLP switch~\cite{hahn2021no} is only at 82.1\%. Also, \ourmethod  exploit$\rightarrow$explore switch is 84.1\% accurate while NRNS doesn't have such a recovery switch (details of this study in Appendix~\ref{subsec:supp-switchExperiment}).

\noindent\textbf{\jw{Large} potential for last-mile navigation.} When \oracledet and \oraclept are used there is a 10.2\% overall improvement in success from 43.5 to 53.7\% (\tabref{tab:ablation-curved}, rows 2 \& 10). Notably, in easy episodes, oracle performance is an ambitious upper bound with an increase in success of 13.9\% (58.7$\rightarrow$72.6\%). 
For the hard (\ie longer) episodes, the oracle components have a relatively lower impact. This is quite intuitive as \explore errors are a more prominent error mode in long-horizon episodes instead of \exploit.

\noindent\textbf{Improvements with \oraclegd.} Even if we assume a perfect variant of \explore~\jw{system} from~\cite{hahn2021no}, we observe that performance saturates at 67.7\% success (row 11, \tabref{tab:ablation-curved}). Comparing rows 11 and 12, \ourmethod can boost this asymptotic success rate by 18.5\% (67.7$\rightarrow$86.2\%).

\noindent\textbf{Analysis: Why is \ourmethod more robust?} In \figref{fig:data-bias}, we visualize the frequency distribution of heading (from the agent to the target) in expert demonstrations~\cite{hahn2021no} (`train GT') and 
that observed at inference (`test GT').
With no geometric structure, NRNS picks up the bias in training data, particularly, towards the heading of 0 (optimal trajectories entail mostly moving forward).
Concretely, 72.2\% of the training data is within $[-15\degree,15\degree]$. This drops to 39.4\% at test time when the \exploit phase is reached (using the best-performing \oraclegd).
Quantified with (first) Wasserstein distance, $W(\text{Test GT},\text{NRNS})=0.0134$ \vs $W(\text{Test GT},\text{\ourmethod})=0.0034$, demonstrating \ourmethod can better match the distribution at inference.
\vspace{-2mm}
\subsection{Physical Robot Experiments}
\begin{figure}[t]
    \centering
    {
        \phantomsubcaption\label{fig:data-bias}
        \phantomsubcaption\label{fig:robotTopDown}
    }
    \setlength{\tabcolsep}{7pt}
    \begin{tabular}{cc}
        \includegraphics[height=0.215\textwidth]{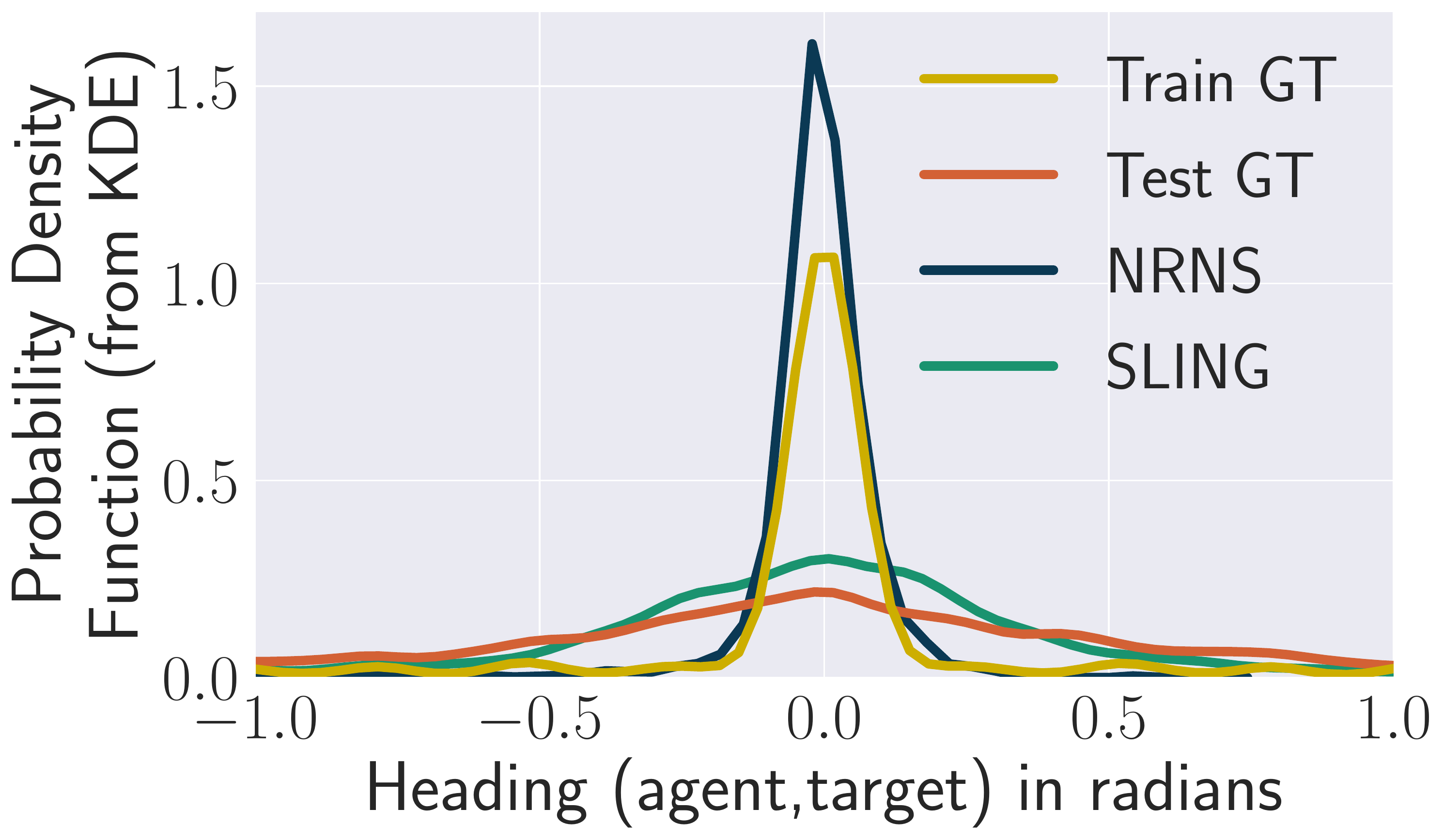}
        &
        \includegraphics[height=0.215\textwidth]{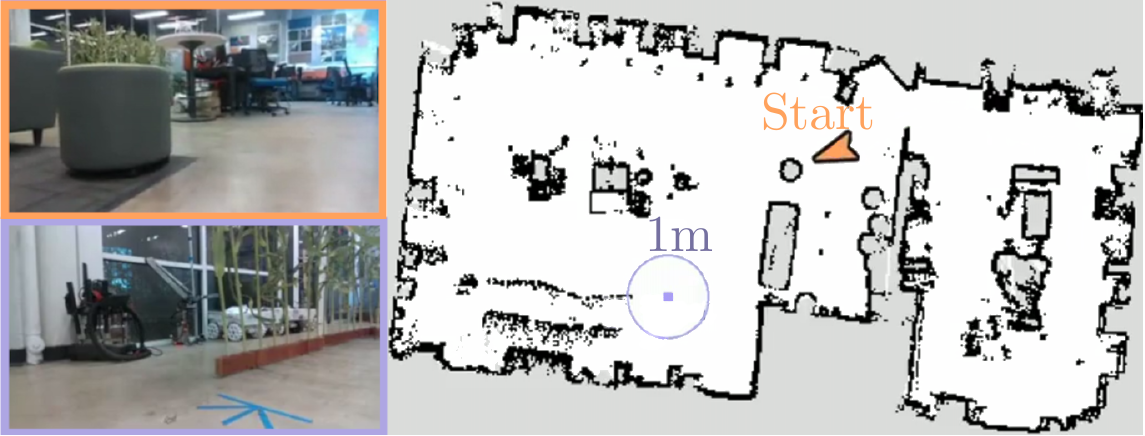}
        \\
        (a) Bias in heading distribution & \jw{(b) Robot experiments (more visuals in Figures~\ref{fig:office-all},~\ref{fig:depart1-all},~\ref{fig:depart2-all})}
    \end{tabular}
    \vspace{2mm}
    \caption{
    (a) Significant distribution shift between training and test heading from agent to goal (\secref{subsec:exp_results}), (b) Navigation policies deployed on a robot in cluttered real-world scenes (\secref{subsec:robot-results}).
    }
    \vspace{-1mm}
\end{figure}

\vspace{-2mm}
\label{subsec:robot-results}
We test the navigation policies on a TerraSentia~\cite{higuti2019under} wheeled robot, equipped with an Intel\textsuperscript{\textregistered} RealSense\textsuperscript{\texttrademark} D435i depth camera (further hardware details in Appendix~\ref{subsec:supp-terrasentia}).
The robot is initialized in an unseen indoor environment and provided an RGB image-goal. We ran a total of 120 trajectories, requiring 30 human hours of effort, across three scenes and two levels of difficulty. Following the previously collected simulation dataset~\cite{hahn2021no}, easy goals are 1.5-3m from the starting location and hard goals are 5-10m from the starting location. Particularly, we test within an office and the common areas in two department buildings, over easy and hard episodes (following definitions from~\cite{hahn2021no}). The physical setup (office) is shown in \figref{fig:robotTopDown}. As in simulation, the agent is successful if it executes \texttt{stop} action within 1m of the goal.
Examples of the image-goal utilized in physical robot experiments and precautions taken are included in Appendix~\ref{subsec:supp-terrasentia}.

As shown in~\tabref{tab:cslrealexp}, for sim-to-real experiments, we base the \explore~\jw{system} on the NRNS model. We choose NRNS as the authors published an instantiation trained exclusively on real-world trajectories, particularly, RealEstate10K~\cite{zhou2018stereo} (house tours videos from YouTube).
\begin{wraptable}{r}{0.4\textwidth}
    \centering
    \vspace{-4mm}
    \resizebox{\linewidth}{!}
        {
        \begin{tabular}{lcccc}
        \toprule
             & \multicolumn{2}{c}{Easy} & \multicolumn{2}{c}{Hard} \\
            Method & 
            Succ$\uparrow$ & 
            SPL$\uparrow$ &
            Succ$\uparrow$ & 
            SPL$\uparrow$
            \\
            \midrule
            \jw{NRNS}~\cite{hahn2021no}& 40.0 &  37.7
            & 3.3 & 3.3
            \\
            +~\ourmethod & \textbf{56.6} & \textbf{53.7}
            &  \textbf{20.0} & \textbf{19.3} 
            \\
        \bottomrule
        \end{tabular}
        }
    \caption{Results in real-world scenes.}
    \label{tab:cslrealexp}
    \vspace{-4mm}
\end{wraptable}
In preliminary experiments, we verified that this NRNS instantiation outperformed its simulation counterpart. For a direct comparison, in \ourmethod + \nrnsgd, we utilize the same \explore~\jw{system} but add our switching and \exploit~\jw{system} (\ourmethod) around it. With \ourmethod, we improve performance from 40.0\% success to 56.6\%. The gains become more prominent as the task horizon increases, leading to an improvement in success rate from 3.3\% to 20.0\%. The large gains in hard episodes (which are exploration heavy) are accounted to \ourmethod's better explore$\rightarrow$exploit switch and \ourmethod's last-mile navigation \jw{system} that is not biased to zero heading (particularly important for curved and long episodes).

\vspace{-2mm}
\section{Conclusion}
\label{sec:conclusion}
\vspace{-2mm}

In this work, we identify and leverage the geometric structure of last-mile navigation for the challenging image-goal navigation task~\cite{hahn2021no}. 
With analysis of data distributions, we demonstrate that learning from expert demonstrations may lead to developing a bias.
Being entirely complementary to prior work, we demonstrate that adding \ourmethod leads to improvements across data splits, episode complexity, and \explore policies, establishing the new state-of-the-art for image-goal navigation~\cite{hahn2021no}.
We also transfer policies trained in simulation to real-world scenes and demonstrate significant gains in performance. 
Further improvements in the switching mechanism, neural keypoint features, visual representations from view augmentations, \etc complement our proposed approach to help improve performance in future work.

Like any method, \ourmethod has several aspects where follow-up works can improve on. We list them explicitly: (1) Our method is limited by mistakes in matching correspondences. (2) We add additional parameters that need to be tuned. (3) We make a single prediction for last-mile navigation. (4) We assume access to depth and pose information. More details of these aspect as well as a discussion on pose errors, depth noise, and the nuanced image-goal navigation definition in Appendix~\ref{subsec:supplimits}.

\clearpage
\acknowledgments{\jw{
We thank the reviewers for suggesting additional experiments to make the work stronger. JW and GC are supported by ONR MURI N00014-19-1-2373. We are grateful to Akihiro Higuti and Mateus Valverde for physical robot help, Dhruv Batra for helping broaden the scope, Jae Yong Lee for help with geometric vision formulation, Meera Hahn for assistance in reproducing NRNS results, and Shubham Tulsiani for helping ground the work better to 3D vision. A big thanks to our friends who gave feedback to improve the submission draft --
Homanga Bharadhwaj,
Raunaq Bhirangi,
Xiaoming Zhao, and
Zhenggang Tang,
}}

\bibliography{corl_main}
\clearpage
\appendix
\section*{Appendix -- \thetitle}
\label{sec:supp}
In this Appendix, we include additional details about the following:
\begin{itemize}\compresslist
    \item [\ref{subsec:supp-implement}] Implementations details of \ourmethod and sensors used for experiments in AI Habitat.
    \item [\ref{subsec:supp-explore}] \Explore modules used for exploration in \ourmethod as well as Image-Goal Navigation Solvers.
    \item [\ref{subsec:supp-straightDataset}] Extended results on the Gibson-straight and MP3D-straight dataset folds.
    \item [\ref{subsec:supp-switchExperiment}] Experimental setup used to compare \ourmethod's geometric switch with \jw{a neural based switch in} Hahn~\etal~\cite{hahn2021no}.
    \item [\ref{subsec:supp-terrasentia}] Hardware configuration and visualizations of real-robot experiments.
    \item [\ref{subsec:supp-curved}] Further results on the curved data folds, ablations, and across multiple seeds.
    \item [\ref{subsec:dtoGoal}] Final distance to goal for top-performing baselines.
    \item [\ref{subsec:stop}] `Stop' budget to show the potential of \exploit (following~\cite{wani2020multion}).
    \item [\ref{subsec:pan}] SLING applied to best prior methods operating on panoramic images.
    \item [\ref{subsec:supplimits}]\jw{ Further analysis of the limitations of SLING.}
\end{itemize}

Code of \ourmethod with \ddppogd and \ovrlgd:\\
\faGithub~\url{https://github.com/Jbwasse2/SLING}

Code of \ourmethod with \straightgd, \nrnsgd, and \oraclegd:\\
\faGithub~\url{https://github.com/Jbwasse2/SLING/tree/nrns}

\section{Implementation Details of \ourmethod}
\label{subsec:supp-implement}

Here we include finer details of \ourmethod that we deferred from the main paper. It was straightforward to connect prior baselines with \ourmethod ($\sim100$ lines of additional code) as prior baselines are reused largely. We were able to test \ourmethod on the robot without any online fine-tuning in the real world.

\noindent\textbf{Input configurations (Extending~\secref{sec:data_and_eval})}
At every time step the agent receives an RGB image, depth image, and the pose of the agent. Policies that are based on \nrnsgd and \straightgd, following prior work~\cite{hahn2021no}, the images are given as $640\times480$ with a FoV of $120\degree$. Policies that are based on \ovrlgd and \ddppogd, for a head-on comparison, the agent is given $128\times128$ images with a FoV of $90\degree$. The pose is given as the position and heading of the agent in the environment.

\noindent\textbf{Depth noise.}
We use the Redwood depth noise model~\cite{Choi_2015_CVPR} to insert noise into the depth image. For the pose noise, we follow the convention from prior work~\cite{hahn2021no,chaplotNeuralTopologicalSLAM2020}, for a direct and fair comparison. 

\noindent\textbf{Pose noise.}
Prior work~\cite{Chaplot2020Explore} built a Gaussian mixture model to capture pose noise from a real-world LoCoBot. We take the same error model, sample from it, and add the sampled noise to the agent's pose.

\noindent\textbf{Local policy (Extending~\secref{subsec:LMNav})}
For a fair comparison, we use the same local policy as prior visual navigation works~\cite{hahn2021no,Chaplot2020Explore}. The local policy takes distance and heading to create a waypoint to navigate to. Building over (near-solved) setting of point-goal navigation, and a fast marching method to build a local map, the agent can localize itself and the goal and navigate towards it. 

\noindent\textbf{Additional hyperparameters.} Beyond the hyperparameters of \explore modules (\straightgd~\cite{chen2018learning}, \ddppogd~\cite{wijmans2019dd,yadav2022OVRL}, \nrnsgd~\cite{hahn2021no}, \ovrlgd~\cite{yadav2022OVRL}, and \oraclegd), our \exploit module introduces only a few hyperparameters which we include in~\tabref{tab:hyperparams-onav}. Note that we use different \# of matches (50 in \ourmethod~+~\nrnsgd \vs 20 in \ourmethod~+~\ovrlgd) because the different methods use different input image sizes. No automated or grid-search tuning was conducted to find these hyperparameters.
\begin{table}[h]
\vspace{-4mm}
    \centering
     \caption{Key hyperparameter choices for \ourmethod.
     } 
    \resizebox{0.8\textwidth}{!}{%
    \setlength{\tabcolsep}{10pt}
    \begin{tabular}{lcc}
        \toprule
        \textbf{Hyperparamter} &  \multicolumn{2}{c}{\textbf{Value}}\\
                \midrule
                \multicolumn{3}{c}{\textit{\exploit module}}\\
        \midrule
        Min \# of Matches for \textit{explore$\rightarrow$exploit} switch (for NRNS/Straight) & \multicolumn{2}{c}{50}\\
        Min \# of Matches for \textit{for explore$\rightarrow$exploit} switch (OVRL) & \multicolumn{2}{c}{20}\\
        Max Predicted Distance & \multicolumn{2}{c}{4 meters}\\
        Confidence threshold for feature matcher module~\cite{sarlin2020superglue}  & \multicolumn{2}{c}{0.5}\\
        
        \bottomrule
    \end{tabular}
    }
    \label{tab:hyperparams-onav}
\end{table}

\section{{\Explore Systems and Image-Goal Navigation Solvers  (Extending~\texorpdfstring{\secref{subsec:baselines}}{}} and \texorpdfstring{\secref{subsec:explore})}{}}
\label{subsec:supp-explore}
The \explore modules, that show the compatibility and efficacy of \ourmethod, are utilized in~\secref{sec:exp}. Next, we include additional details for these.

\jw{
\noindent\textbf{Behavior Cloning with Spatial Memory.} Applies imitation learning (IL) wherein $\agentimage$ and $\goalimage$ are represented with ResNet18~\cite{he2016deep}. Moreover, using the depth map $\agentdepth$, the observations are represented as a spatial metric map (found to be effective across embodied AI tasks~\cite{Chaplot2020Explore,wani2020multion}). 

\noindent\textbf{Behavior Cloning with GRU.} Another IL baseline where the observations at each time step are encoded identically to the above. However, instead of the spatial metric map, a GRU~\cite{ChoARXIV2015} is employed (CNN-RNN architectures have been effective for semantic navigation~\cite{GuptaCVPR2017,gupta2017unifying,chen2019audio,ZhuARXIV2017}).

\noindent\textbf{Zero Experience Replay (ZER)~\cite{al2022ZER}}. A recent plug-and-play RL policy trained using rewards obtained from moving closer to the goal and looking towards it, as well as view augmentation.
The authors shared metrics over the folds of the benchmark~\cite{hahn2021no}, particularly Gibson-curved and \textit{cross-domain} transfer results on MP3D, which we include in~\tabref{tab:main-curved}.

\noindent\textbf{DDPPO~\cite{wijmans2019dd}, NRNS~\cite{hahn2021no}, OVRL~\cite{yadav2022OVRL}.} These methods have been described as part of goal discovery, see~\secref{subsec:explore}. For DDPPO, we report results from Hahn~\etal~\cite{hahn2021no}, trained for 100M steps (10x more compute than NRNS). For NRNS, we report the reproducible metrics from their
\href{https://github.com/meera1hahn/NRNS\#scores}{official implementation} (differs slightly from the paper~\cite{hahn2021no}). At the time of submission, OVRL is the best-performing method on Gibson-curved fold of the benchmark~\cite{hahn2021no}. For OVRL, we requested their checkpoints and re-evaluated them to report detailed metrics across easy-medium-hard folds.

\jw{\noindent\textbf{DDPPO-LMN} }
DDPPO~\cite{wijmans2019dd} is a widely-adopted end-to-end deep RL baseline for embodied AI tasks~\cite{ye2020auxiliary,ye2021auxiliary,yadav2022OVRL} in AIHabitat~\cite{habitat19iccv}. We train DDPPO, exclusively for last-mile navigation. This last-mile navigation DDPPO (termed DDPPO-LMN) was trained on agents initialized at most 3m from the goal. DDPPO-LMN was trained to convergence on these trajectories over 400M steps, in the Gibson scenes. For a fair comparison, we allow DDPPO-LMN to use the same explore$\rightarrow$exploit switch as SLING.}

\noindent\textbf{\straightgd}. Following the strategies of robot vacuums and studies in~\cite{chen2018learning}, this exploration module moves straight till it collides with an obstacle and then turns right ($15^{\circ}$). A collision is estimated if, after completing an action, the pose difference between the agent's movement and its expected displacement is less than $0.1m$.

\noindent\textbf{\nrnsgd}. As introduced by Hahn~\etal~\cite{hahn2021no}, NRNS utilizes four graph convolution layers to extract an embedding from the topological map. The extracted graph embedding (of size 768) along with the goal embedding are fed into a linear layer which predicts the distance estimate from the unexplored nodes to the goal. The next node that the agent navigates to is the node that minimizes this distance plus the distance from the agent to the node. We remove redundant nodes from being added to the topological map, which led to improved performance.

\noindent\textbf{\ddppogd}. For \ddppogd, we used the trained checkpoint we obtained from the OVRL~\cite{yadav2022OVRL} authors, added SLING over it and evaluated it on different data folds. The DDPPO agent was trained for 500M (NRNS~\cite{hahn2021no} train for a max of 100M) steps over RGB observations on the episode dataset from ~\cite{mezghani2021memory}.

\noindent\textbf{\ovrlgd}. For our OVRL experiments, we use the pretrained visual encoders provided by OVRL's authors. The downstream policy is trained using DDPPO on either the MP3D episode dataset (matching~\cite{hahn2021no}) or the Gibson episode dataset (following~\cite{mezghani2021memory}) for their respective experiments. We match OVRL's~\cite{yadav2022OVRL} training setup by using a set of 32 GPUs with 10 episodes each and train the agent for 500M steps. Each worker is allowed to collect up to 64 frames of experience in the environment and then trained using 2 PPO epochs with 2 mini-batches, with a learning rate of 2.5 x 10$^{-4}$.

\noindent\textbf{\oraclegd}. This has the same architecture as NRNS-GD from Hahn~\etal~\cite{hahn2021no}, where the agent builds a topological map to represent the environment. However, this method has privileged access, particularly, to the perfect distances from each node in the map to the goal. The planner will then deterministically navigate the agent to the node in the map that has the lowest distance to the goal. We also found additional tweaks and edits to improve performance: (1) removing the agent to node distance and (2) removal of redundant nodes in the topological map. 

Importantly, even when using \oraclegd for \explore, the \exploit modules do not have access to the ground truth distance to the goal. Furthermore, the oracle does not use its information to switch between \explore and \exploit. So there is still a long way to perfect navigation, despite using a \oraclegd.

\section{Results on Straight Data Folds (Extending~\texorpdfstring{\secref{subsec:exp_results}}{})}
\label{subsec:supp-straightDataset}
In~\tabref{tab:main-straight}, we supplement the results of Gibson-curved and MP3D-curved (from~\tabref{tab:main-curved} and ~\tabref{tab:indomain-ovrl}), to include takeaways based on the straight counterparts. ZER~\cite{al2022ZER} does not report results on the straight split, hence, could not be included in~\tabref{tab:main-straight}.

\begin{table}[h!]
\centering
\caption{\textbf{Results for `Gibson-straight' and `MP3D-straight' episodes.} Note the significant gains by adding \ourmethod to prior works. \ourmethod~+~\nrnsgd performs the best on Gibson-straight and \ourmethod~+~\straightgd performs best on  MP3D-straight.}
\resizebox{\textwidth}{!}{
\begin{tabular}{l@{\hskip 4mm} l c c c c c c c c}
    \toprule
    &&\multicolumn{2}{c}{Overall}&\multicolumn{2}{c}{Easy}&\multicolumn{2}{c}{Medium}&\multicolumn{2}{c}{Hard}\\
    &Method & Succ$\uparrow$ & SPL$\uparrow$ & Succ$\uparrow$ & SPL$\uparrow$ & Succ$\uparrow$ & SPL$\uparrow$ & Succ$\uparrow$ & SPL$\uparrow$\\
    \midrule
    \multicolumn{10}{c}{Dataset = \textit{Gibson-straight}}\\
    \midrule
    1& BC w/ Spatial Memory~\cite{bain1995framework} & 12.5 & 12.1 &24.8 & 23.9 & 11.5 & 11.2 & 1.3 & 1.2\\
    2& BC w/ GRU State~\cite{bain1995framework,ChoARXIV2015}& 19.5 & 18.7 & 34.9 & 33.4 & 17.6 & 17.0 & 6.0 & 5.9\\
    3& DDPPO~\cite{wijmans2019dd} (from~\cite{hahn2021no}) & 29.0 & 26.8 & 43.2 & 38.5 & 36.4 & 34.8 & 7.4 & 7.2\\
    4 & OVRL~\cite{yadav2022OVRL} & 44.9 &30.0 &53.6& 34.7 & 48.6 & 33.3 & 32.5 & 21.9\\
    5& NRNS~\cite{hahn2021no}& 47.3 & 39.8 & 70.1 & 62.7 & 50.7 & 41.5 & 21.2 & 15.4\\
    6 & \ourmethod + \straightgd & 63.5 & 55.5 & 84.3 & \textbf{79.0} & 65.6 & 57.3 & 40.6 & 30.2 \\
    7 & \ourmethod + \ddppogd & 38.6 & 26.0 & 54.9 & 39.5 & 41.0 & 27.2 & 20.0 & 11.3 \\
    8 & \ourmethod + \ovrlgd & 58.1 & 42.5 & 71.2 & 54.1 & 60.3 &44.4 & 43.0 & 29.1 \\
    9 & \ourmethod + \nrnsgd \textit & \textbf{68.4} & \textbf{58.0}  & \textbf{85.0} & 76.8 & \textbf{71.3} & \textbf{60.6} & \textbf{49.0} & \textbf{36.6}\\
    \midrule
    \multicolumn{10}{c}{Dataset = \textit{MP3D-straight}}\\
    \midrule
    10& \jw{BC w/ Spatial Memory}~\cite{bain1995framework} & 13.3 & 12.7 & 25.8 & 24.8 & 11.3 & 10.6 & 3.0 & 2.9\\
    11& \jw{BC w/ GRU State}~\cite{bain1995framework,ChoARXIV2015}& 15.7 & 15.4 & 30.2 & 29.5 & 12.7 & 12.4 & 4.4 & 4.3\\
    12& DDPPO~\cite{wijmans2019dd} (from~\cite{hahn2021no}) & 27.4 & 24.5 & 36.4 & 30.8 & 33.8 & 31.4 & 12.0 & 11.5\\
    13& OVRL~\cite{yadav2022OVRL} & 52.6 & 39.4 & 69.5 & 54.0 & 51.7 & 39.2 & 36.7 & 25.0\\
    14& NRNS~\cite{hahn2021no} &36.7&30.2& 56.9 & 49.2 & 33.7 & 27.1 & 19.6 & 14.4\\
    15 & \jw{\ourmethod + \straightgd} & \textbf{61.3} & \textbf{54.3} & \textbf{83.0} & \textbf{77.9} & \textbf{60.2} & \textbf{52.3} & \textbf{40.9} & \textbf{32.9}\\
    16 & \ourmethod + \ddppogd \tablefootnote{Due to limited compute, we were unable to retrain DDPPO-GD from scratch. Therefore, we use DDPPO-GD trained on Gibson, without SLING this model had an overall success and SPL of 9.0\% and 4.4\% respectively.} & 31.7 & 21.4 & 49.6 & 36.2 & 31.4 & 20.1 & 14.2 & 7.9\\
    17 & \ourmethod + \ovrlgd & 58.3 & 47.1 & 78.8 & 68.5 & 58.7 & 46.3 & 37.4 & 26.5 \\
    18& \ourmethod + \nrnsgd & 60.6 &49.6 &82.0 & 76.1 & 59.1 & 46.5 & 40.8 & 26.3 \\
    \bottomrule
\end{tabular}
}
\label{tab:main-straight}
\end{table}

\noindent\textbf{State-of-the-art performance also on straight data folds.} Similar to the curved results, utilizing \ourmethod with previous \explore modules results in the highest performance across the Gibson-straight and MP3D-straight data folds. Over previous state-of-the-art (NRNS~\cite{hahn2021no}), we improve the success rate by 21.1\% (rows 5 \& 9 under overall success) on Gibson and by \jw{24.6\%} (rows 14 \& \jw{15} under overall success) on the MP3D dataset. 

\noindent\textbf{\ourmethod significantly boosts all prior policies.} On the Gibson dataset, using \ourmethod improved the success rate on NRNS by 21.1\% (rows 5 and 9), on DDPPO by 9.6\% (rows 3 and 7), and OVRL by 13.2\%  (rows 4 and 8). Similar trends hold for the MP3D dataset. Quite surprisingly, even using the very simple \straightgd with \ourmethod (row 6) works really well for the straight fold. \jw{Note that this straight-exploring agent outperforms all other neural policies on MP3D.} 

\section{Details of Switch Experiment (Extending~\texorpdfstring{\secref{subsec:exp_results}}{})}
\label{subsec:supp-switchExperiment}
In~\secref{subsec:exp_results}, under `Geometric switches are better', we presented that our switches are more accurate.
Particularly, \ourmethod's explore$\rightarrow$exploit switch is 92.0\% accurate and MLP switch~\cite{hahn2021no} is only at 82.1\%. Also, \ourmethod  exploit$\rightarrow$explore switch is 84.1\% accurate while NRNS doesn't have such a recovery switch. These are summarized in~\tabref{tab:switchResults}.
Next, we provide the deferred details of this study and evaluation data.

\begin{figure}[h]
    \centering
    {
        \phantomsubcaption\label{fig:simpair}
        \phantomsubcaption\label{fig:notsimpair}
    }
    \setlength{\tabcolsep}{7pt}
    \begin{tabular}{cc}
        \includegraphics[height=0.17\textwidth]{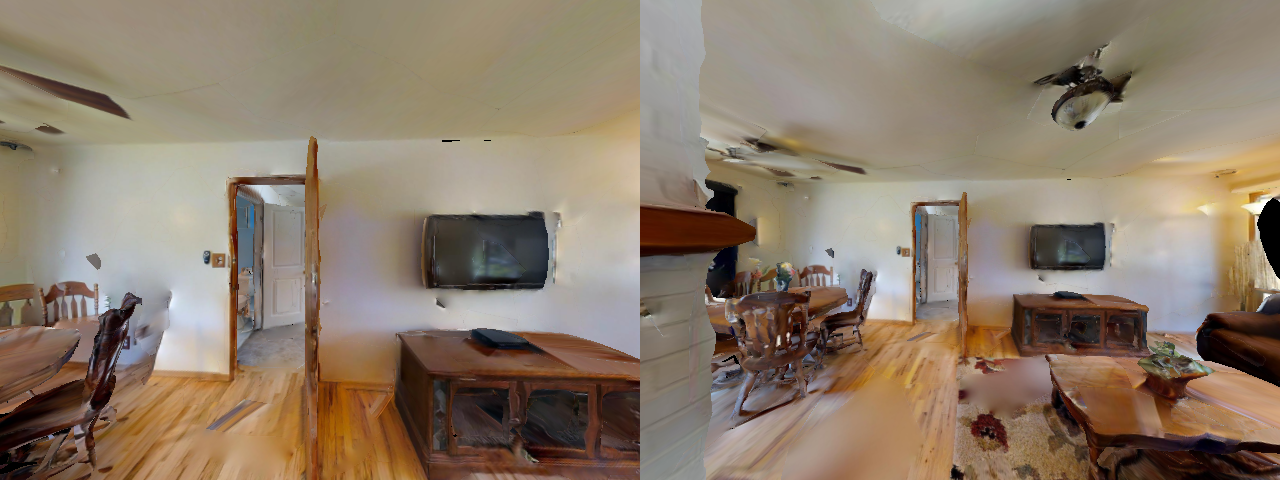}
        &
        \includegraphics[height=0.17\textwidth]{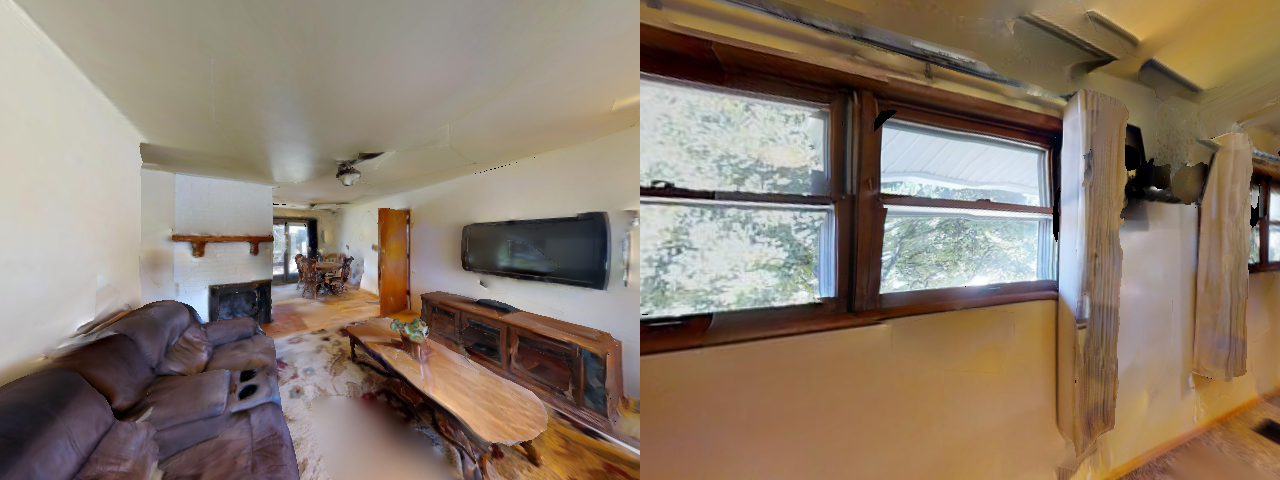}
        \\
        (a) Positive image pair & (b) Negative image pair
    \end{tabular}
    \vspace{-2mm}
    \caption{\textbf{Positive and negative pairs for \exploit.}
    (a) The given image pair is similar (or positive) as the views are close and have significant overlap. (b) The image pair is dissimilar (or a negative) because they were taken in different rooms (their euclidean distance and geodesic to euclidean distance ratio are quite high ($>1.2$).
    }
\end{figure}

We sampled 500 image pairs per scene -- 250 positives and 250 negatives for last-mile navigation. These are sampled randomly from 13 test environments (a total of $6500$ image pairs). What is a positive for last-mile navigation?
This is not well defined in prior works~\cite{hahn2021no,chaplotNeuralTopologicalSLAM2020,meng2020scaling}.
We say two views are positive (for \exploit), if they are (1) less than $3m$ apart, (2) the angular difference is less than $22.5\degree$, and (3) the ratio of the geodesic distance over the euclidean distance is less than 1.2. 
An example of a positive and negative pair for \exploit is visualized in~\figref{fig:simpair} and~\figref{fig:notsimpair}.

\begin{table}[h!]
\vspace{-3mm}
    \centering
    \caption{\textbf{Comparing accuracy of switches.} Our \textit{explore$\rightarrow$exploit} simple switching mechanisms are more accurate than MLP switches~\cite{hahn2021no}.}
    \begin{tabular}{ccc}
    \toprule
        Switching Mechanism & \textit{Explore$\rightarrow$Exploit} Accuracy & \textit{Exploit$\rightarrow$Explore} Accuracy\\ \midrule
        MLP switch from NRNS~\cite{hahn2021no} & 82.1 & N/A\\
        \ourmethod switch & \textbf{92.0} & 84.1\\
    \bottomrule
    \end{tabular}
    
     \label{tab:switchResults}
\end{table}

\section{Robotic Experiments (Extending~\texorpdfstring{\secref{subsec:robot-results}}{})}
\label{subsec:supp-terrasentia}
While we included all major real-robot results in the main paper (\secref{subsec:robot-results}), we deferred several details, which we describe next.

\noindent\textbf{Sensing details.}
The TerraSentia robot utilizes an Intel\textsuperscript{\textregistered} RealSense\textsuperscript{\texttrademark} D435i depth camera with a horizontal and vertical FoV of $69\degree$ and $42\degree$, respectively. The depth image is spatially aligned to the RGB image. When using Robot Operating System (ROS), the RGB and depth images are not necessarily published at the same time. Therefore, RGB and depth images are paired with the closest temporal message. To obtain the pose estimate, we utilize ORB-SLAM2~\cite{murORB2}. 
\begin{figure}[t]
    \centering
    {
        \phantomsubcaption\label{fig:labtop}
        \phantomsubcaption\label{fig:labphoto}
        \phantomsubcaption\label{fig:labgoals}
    }
        \begin{tabular}{cc}
    \includegraphics[width=0.95\textwidth]{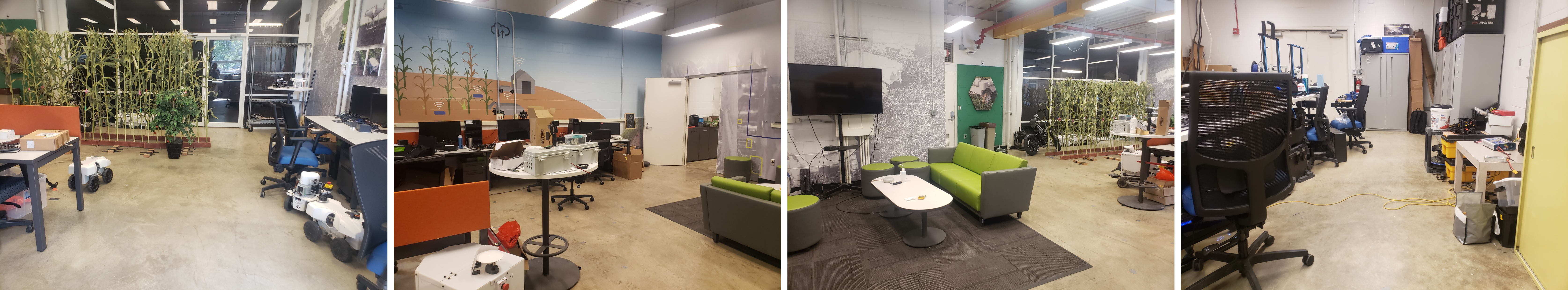}\\
        (a) Scene snapshots
    \end{tabular}
    \includegraphics[height=0.563\textwidth]{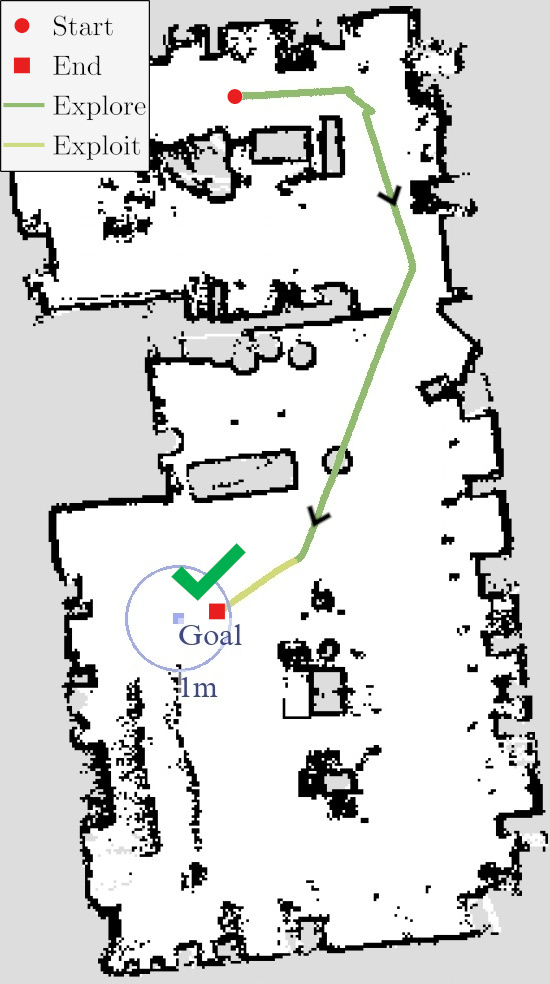}
    \includegraphics[height=0.563\textwidth]{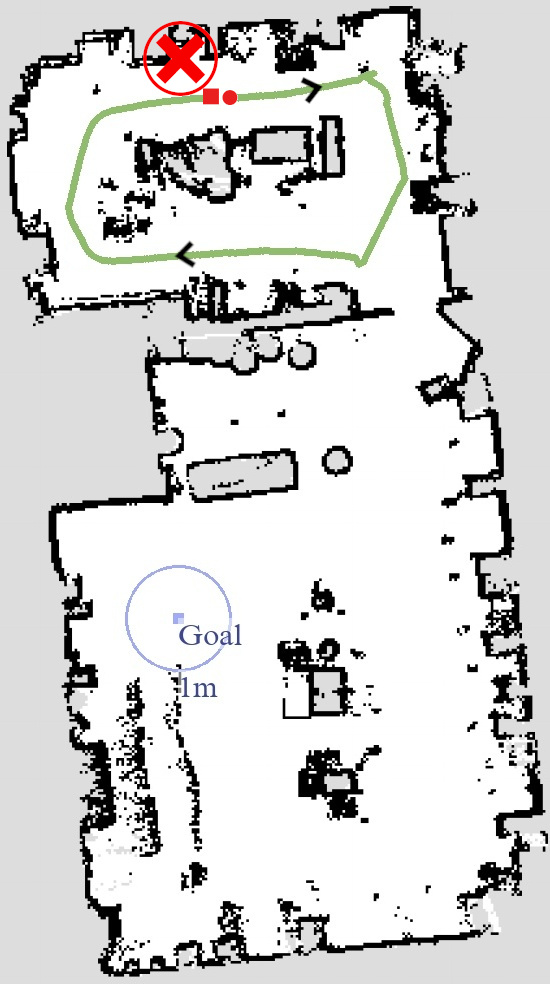}
    \includegraphics[height=0.563\textwidth]{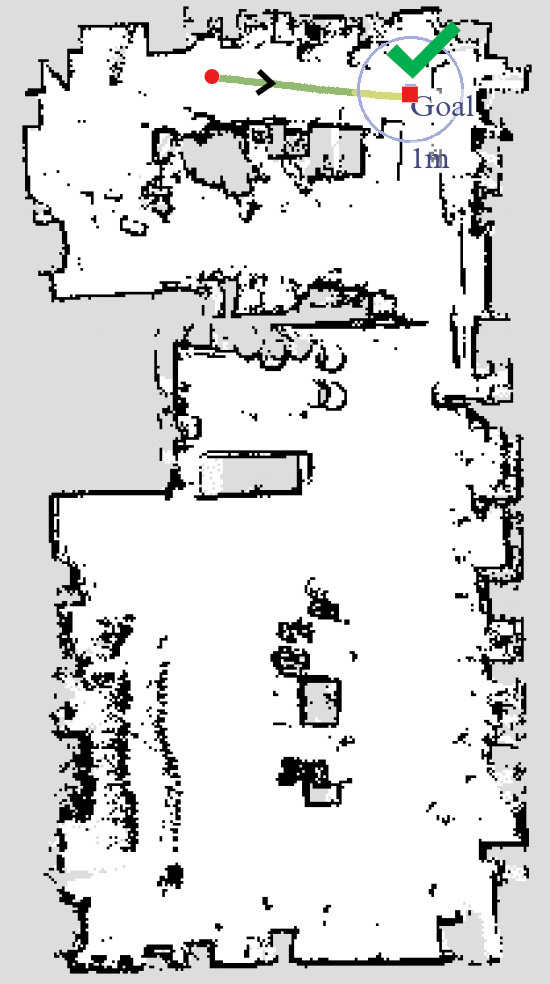}\\ (b) SLING example trajectories. \\
    \includegraphics[height=0.563\textwidth]{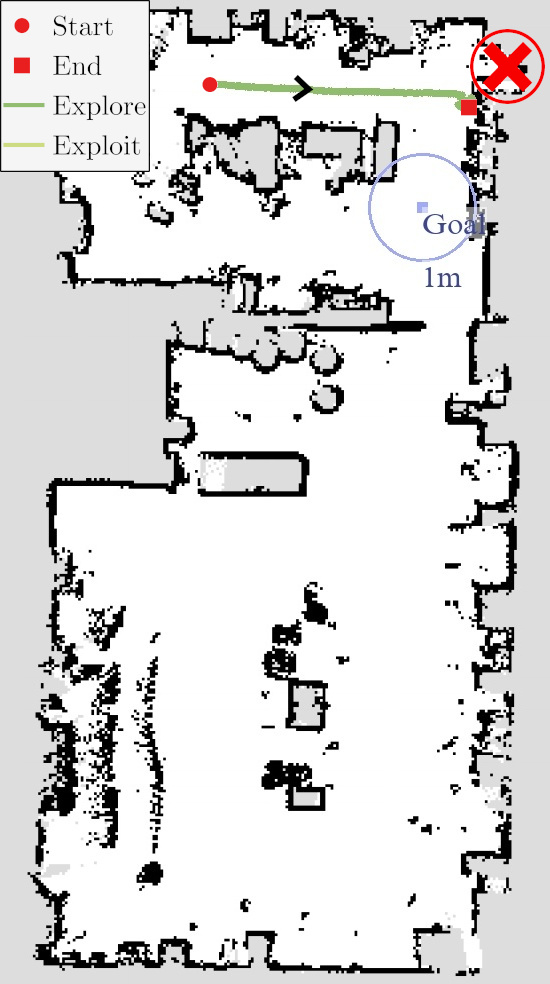}
    \includegraphics[height=0.563\textwidth]{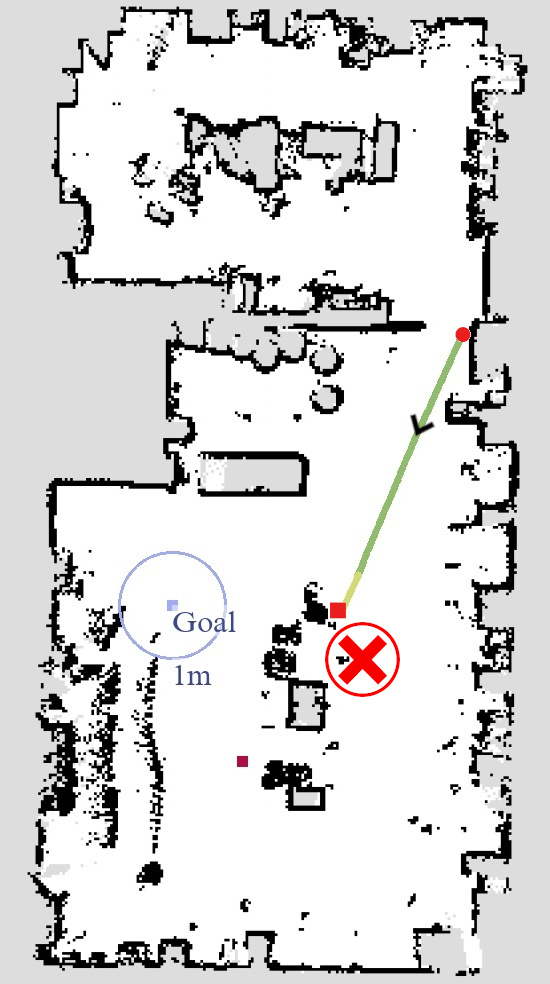}
    \includegraphics[height=0.563\textwidth]{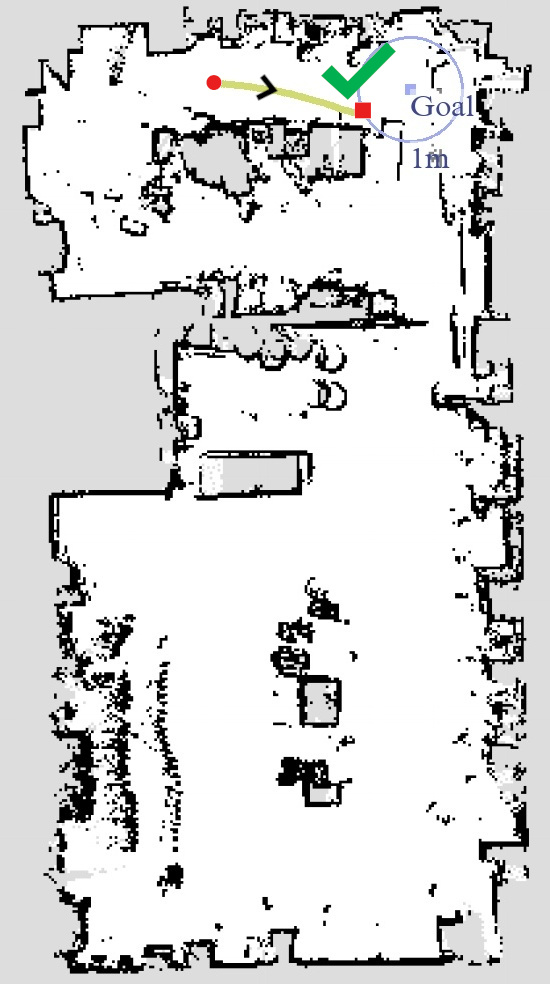}\\
    (c) NRNS example trajectories.
     \caption{\textbf{\texttt{\jw{Office} scene}.} The map size is approximately $30m$ by $11m$.
    (a) office equipment serves as obstacles in this scene (b,c) Topdown map reconstruction with  \href{http://introlab.github.io/rtabmap/}{RTAB-Map} of image-goal navigation task.
    }
    \label{fig:office-all}
\end{figure}

\noindent\textbf{Safety.} In order to protect the motors on the robot from getting damaged, and to make the image-goal task more realistic, we stop the robot when it crashes into an obstacle. After stopping we take the measurements needed to acquire the reported metrics. %

\noindent\textbf{Nonlinear model predictive control.} Our real-robot system deploys Nonlinear Model Predictive Control (NMPC)~\cite{gasparino2022wayfast} for the robot to execute actions. We utilize skid-steer dynamics to model the behavior of the TerraSentia. The controller optimizes the cost function consisting of penalties for errors between the robot's states and states up to and including the final estimation state, and the magnitude of the control input.

\begin{figure}[t]
    \centering
    {
        \phantomsubcaption\label{fig:halltop}
        \phantomsubcaption\label{fig:hallphoto}
        \phantomsubcaption\label{fig:hallgoals}
    }
        \begin{tabular}{cc}
    \includegraphics[width=0.95\textwidth]{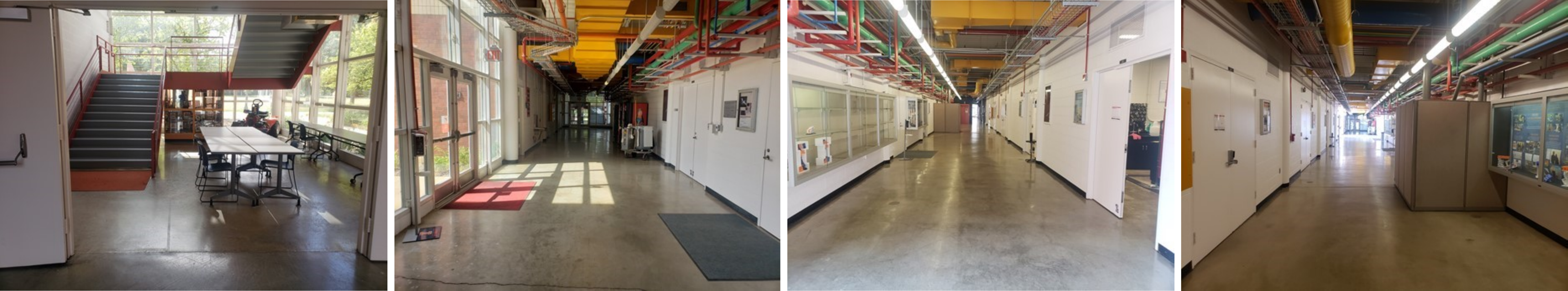}\\
        (a) Scene snapshots
    \end{tabular}
    \includegraphics[height=0.92\textwidth]{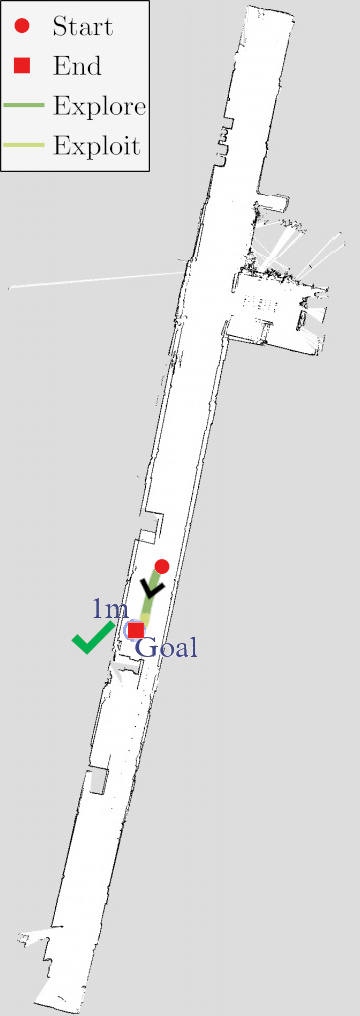}
    \includegraphics[height=0.92\textwidth]{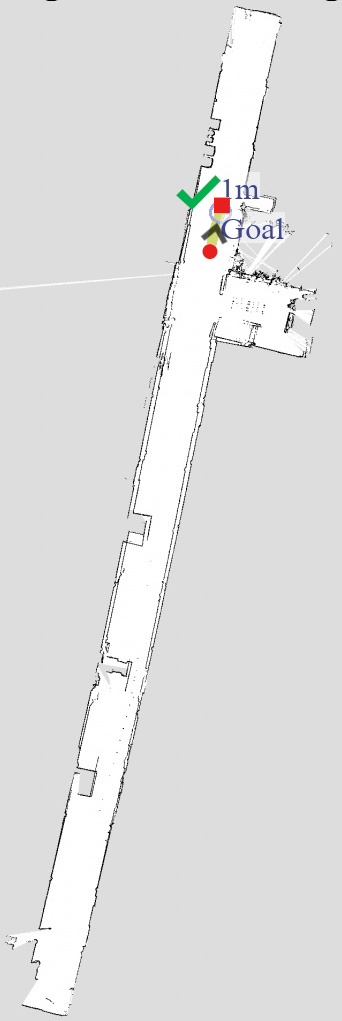}
    \includegraphics[height=0.92\textwidth]{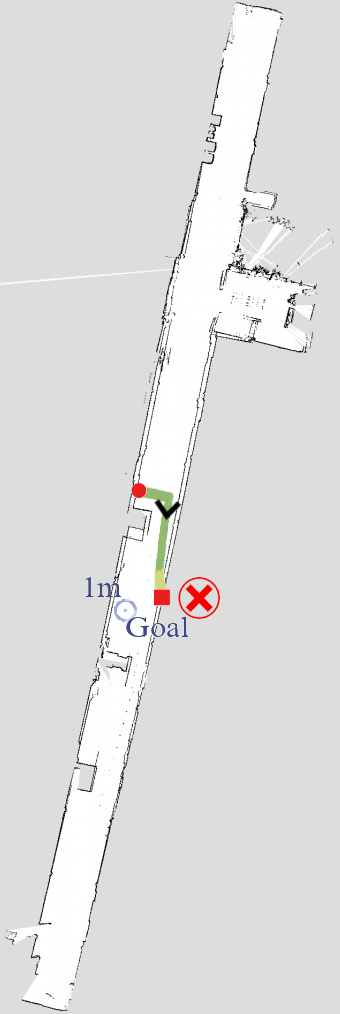}\\
    (b) SLING examples
    \caption{\textbf{\texttt{\jw{Department1}} scene.} The Map size is approximately $95m$ by $32m$ (large width due to the LIDAR going through a door in the upper left side of image).
    (a) long corridors make images agent's views quite similar and make navigation challenging (b) Topdown map reconstruction with  \href{http://introlab.github.io/rtabmap/}{RTAB-Map} of image-goal navigation task.
    }
    \label{fig:depart1-all}
\end{figure}

\begin{figure}[t]
    \centering
    {
        \phantomsubcaption\label{fig:csltop}
        \phantomsubcaption\label{fig:cslphoto}
        \phantomsubcaption\label{fig:cslgoals}
    }
        \begin{tabular}{cc}
    \includegraphics[width=0.95\textwidth]{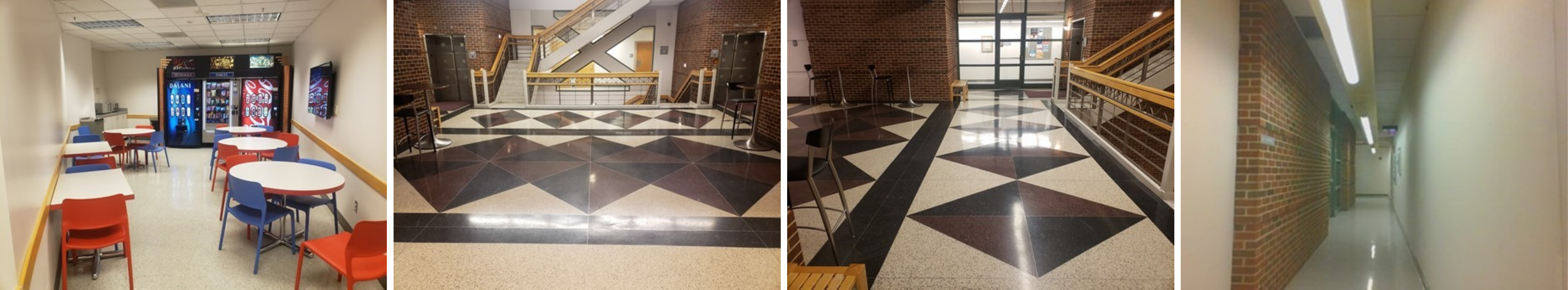}\\
        (a) Scene snapshots
    \end{tabular}
    \includegraphics[height=0.455\textwidth]{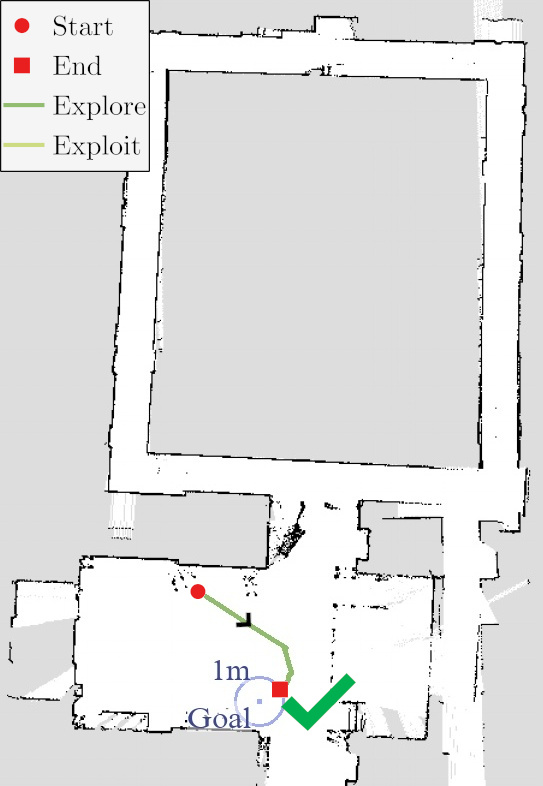}
    \includegraphics[height=0.455\textwidth]{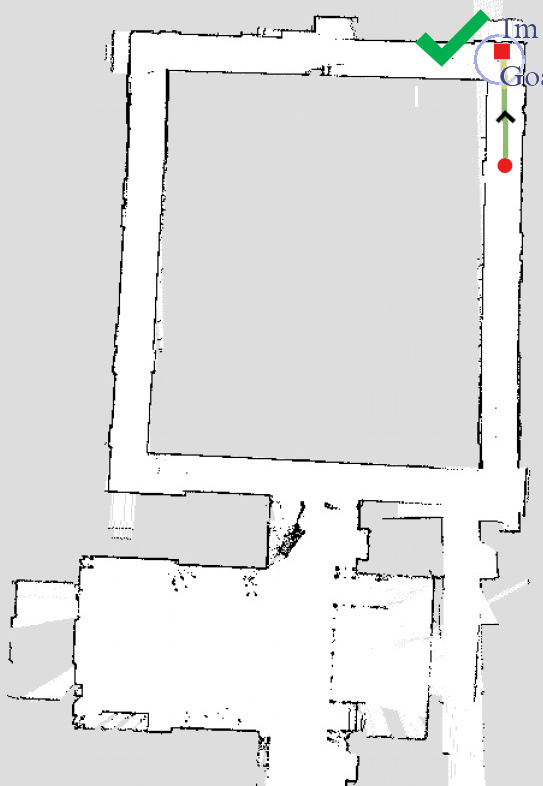}
    \includegraphics[height=0.455\textwidth]{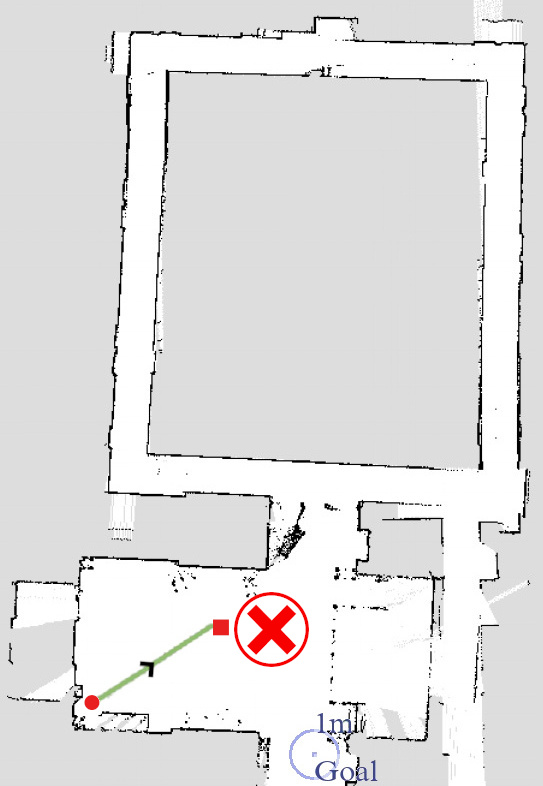}\\ 
    (b) SLING examples\\
    \includegraphics[height=0.455\textwidth]{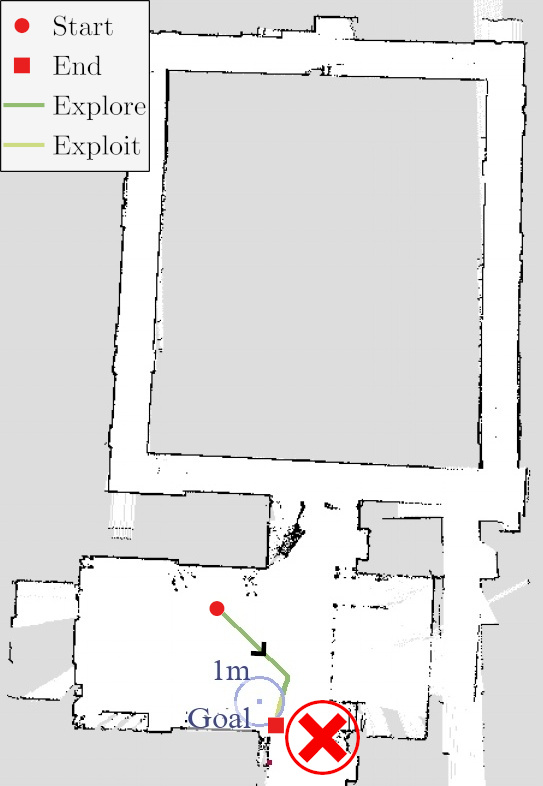}
    \includegraphics[height=0.455\textwidth]{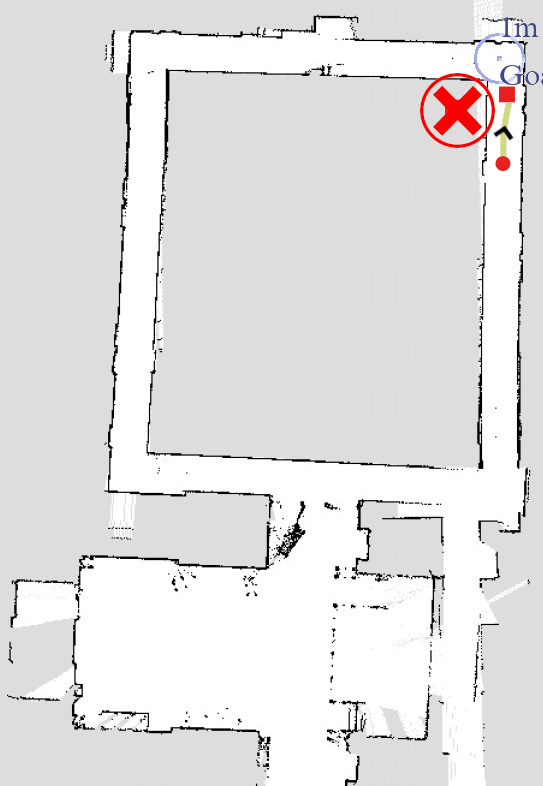}
    \includegraphics[height=0.455\textwidth]{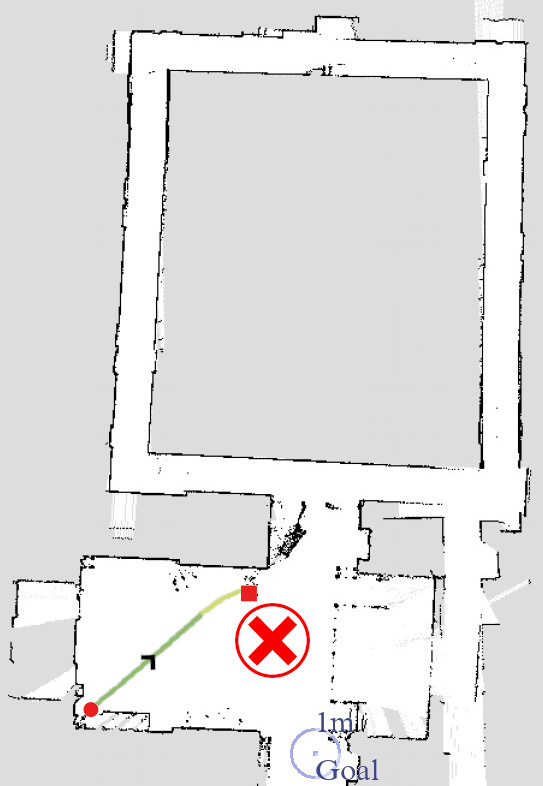}\\
    (c)  NRNS examples.
    \caption{\textbf{\texttt{\jw{Department2}} scene.}  The map size is approximately $42m$ by $22m$ for the shown floor.
    (a) several furniture items and specular floors are challenging for navigation, (b,c) Topdown map reconstruction with  \href{http://introlab.github.io/rtabmap/}{RTAB-Map} of image-goal navigation task.    }
    \label{fig:depart2-all}
\end{figure}

\noindent\textbf{Environments and examples}. We choose a diverse set of three scenes for our real-robot study including a total of 120 demonstrations. These environments are challenging, containing diverse layouts and furniture, several obstacles, varying lighting conditions, long hallways, and visually-confounding common spaces (due to repeated patterns.) We show qualitative examples for each scene, particularly, a reconstruction from the robot, third-person views to show the scene, and trajectory examples. \jw{These trajectories and reconstructions were not used for real-world robotics experiments. They are strictly added for visualization purposes.} The \texttt{office}, \texttt{department1}, and \texttt{department2} environments are visualized in \figref{fig:office-all}, \figref{fig:depart1-all}, and \figref{fig:depart2-all}, respectively.


\section{Additional Results for Gibson-curved and MP3D-curved (Extending~\texorpdfstring{\secref{sec:exp}}{})}
\label{subsec:supp-curved}
We supplement the results on the curved dataset. Particularly, we include ablations of \ourmethod + \ovrlgd~(\tabref{tab:ablation-curved-ovrl}), MP3D curved episodes~(\tabref{tab:indomain-ovrl}), and multi-seed runs~(\tabref{tab:seeds-ovrl-ddppo}).

\noindent\textbf{Ablation results on OVRL (Extending~\tabref{tab:ablation-curved}).} We performed several ablations (see~\secref{subsec:baselines}) on the \ourmethod~+~\nrnsgd method in~\tabref{tab:ablation-curved}. We test the same on the \ourmethod~+~\ovrlgd in \tabref{tab:ablation-curved-ovrl}.
Results follow a trend similar to \ourmethod~+~\nrnsgd, with performance drops when key components of \ourmethod are taken away. The biggest drop (overall success drop from 55.4$\rightarrow$37.9) is observed without our switches \ie replacing our explore$\rightarrow$exploit switch with MLP switch~\cite{hahn2021no} and removing the exploit$\rightarrow$explore switch (useful for error recovery).
Notably, our method shows resistance to noise with an SPL change of +0.2\% (rows 2 and 8) when depth noise is added. Unlike the \nrnsgd counterpart, \ovrlgd \explore module is quite less resilient to pose noise.

\begin{table}[h!]
\centering
\caption{\textbf{\ourmethod + \ovrlgd ablations on `Gibson-curved' episodes.} Ablations demonstrate the need for using two switches as well as utilizing learned features. Further testing demonstrates \ourmethod + \ovrlgd is resilient to sensor noise.}
\resizebox{\textwidth}{!}{
\begin{tabular}{l@{\hskip 4mm} l c c c c c c c c}
    \toprule
    &&\multicolumn{2}{c}{Overall}&\multicolumn{2}{c}{Easy}&\multicolumn{2}{c}{Medium}&\multicolumn{2}{c}{Hard}\\
    &Method & Succ$\uparrow$ & SPL$\uparrow$ & Succ$\uparrow$ & SPL$\uparrow$ & Succ$\uparrow$ & SPL$\uparrow$ & Succ$\uparrow$ & SPL$\uparrow$\\
    
    \midrule
    1& OVRL~\cite{yadav2022OVRL} & 45.6 & 28.0 & 53.6 & 31.7 & 47.6 & 30.2 & 35.6 & 21.9 \\
    \midrule
    2& \ourmethod + \ovrlgd & 54.8 & 37.3 & 65.4 & 45.7 & 59.5 & 40.6 & 39.6 & 25.5\\
    3& w/ MLP Switch & 43.5& 18.0& 50.9& 19.5& 46.0& 21.7& 31.9& 16.3\\
    4& w/ MLP Switch w/o Recovery & 37.9& 16.7& 47.7& 18.1& 43.0& 21.3& 28.2& 15.2\\
    5& w/o Neural Features & 53.7 & 34.6 & 64.0 & 41.9 & 56.9 & 36.8& 40.2 & 25.2\\
    6& w/ Pose Noise & 46.6& 29.2& 55.1& 33.0& 50.4& 33.0& 34.4& 21.5\\
    7& w/ Pose \& Depth Noise & 46.0& 28.4& 54.7& 33.1& 49.5& 31.2& 33.8& 20.9\\
    8& w/ Depth Noise & 55.8& 37.6& 67.6& 45.9& 58.1& 40.3& 41.8& 26.7\\
    \bottomrule
\end{tabular}
}
\label{tab:ablation-curved-ovrl}
\end{table}

\noindent\textbf{Additional in-domain MP3D results (extending~\tabref{tab:main-curved}).} %
\jw{Consistent with previous results, \ourmethod improves performance across several methods (compare rows 3 \vs 7, 4 \vs 8, and 5 \vs 9 of \tabref{tab:indomain-ovrl}). Results for ZER are not included as they do not present results for this split.}

\begin{table}[ht]
\centering
\caption{\jw{\textbf{Results for `MP3D-curved' episodes.} Extending~\tabref{tab:main-curved}, adding \ourmethod to prior works improves navigation results.}}
\resizebox{\textwidth}{!}{
\begin{tabular}{l@{\hskip 4mm} l c c c c c c c c}
    \toprule
    &&\multicolumn{2}{c}{Overall}&\multicolumn{2}{c}{Easy}&\multicolumn{2}{c}{Medium}&\multicolumn{2}{c}{Hard}\\
    &Method & Succ$\uparrow$ & SPL$\uparrow$ & Succ$\uparrow$ & SPL$\uparrow$ & Succ$\uparrow$ & SPL$\uparrow$ & Succ$\uparrow$ & SPL$\uparrow$\\
    \midrule
    \multicolumn{10}{c}{Dataset = \textit{MP3D-curved} }\\
    \midrule
    1& BC w/ Spatial Memory~\cite{bain1995framework} & 2.2 & 1.9 & 4.9 & 4.2 & 1.4 & 1.2 & 0.4 & 0.3\\
    2& BC w/ GRU~\cite{bain1995framework,ChoARXIV2015} & 1.3 & 1.1 & 3.1 & 2.6 & 0.8 & 0.7 & 0.1 & 0.0\\
    3& DDPPO~\cite{wijmans2019dd} (from~\cite{hahn2021no}) & 12.9 & 10.0 & 17.9 & 13.2 & 15.0 & 12.1 & 5.9 & 4.8\\
    4& NRNS~\cite{hahn2021no} &15.5 & 7.4 & 23.1 & 10.8 & 15.1 & 7.3 & 8.4 & 4.1\\
    5& OVRL~\cite{yadav2022OVRL} & 41.6 & 24.4 & 52.4 & 35.2 & 42.6 & 26.3 & \textbf{29.7} & 16.9 \\
    6 & SLING + \straightgd & 25.3 & 10.8 & 31.1 & 11.9 & 27.2 & 12.1 & 17.7 & 8.5\\
    7 & SLING + \ddppogd \tablefootnote{Due to limited compute, we were unable to retrain DDPPO-GD from scratch. Therefore, we use DDPPO-GD trained on Gibson, without SLING this model had an overall success and SPL of 7.5\% and 3.2\% respectively.}& 27.1 & 15.8 & 41.1 & 25.3 & 27.7 & 15.5 & 12.6 & 6.5\\
    8 & \ourmethod + \nrnsgd & 32.6 &14.9 & 43.2 & 19.7 & 32.5 & 15.1 & 22.1 & 9.9\\
    9 & \ourmethod + \ovrlgd& \textbf{46.7} & \textbf{30.1} & \textbf{62.6} & \textbf{41.1} & \textbf{48.4} & \textbf{31.5} & 29.2 & \textbf{17.7}\\
    \bottomrule
\end{tabular}
}
\label{tab:indomain-ovrl}
\end{table}

\noindent\textbf{Multi-seed runs for \ovrlgd and \ddppogd (Extending~\tabref{tab:main-curved}).} In this experiment we ran \ourmethod + \ovrlgd and \ourmethod + \ddppogd on 2 more seeds and present the results in~\tabref{tab:seeds-ovrl-ddppo}. We see a slight improvement in the performance of both of the methods compared to the results presented in~\tabref{tab:main-curved}. Note that Yadav~\etal~\cite{yadav2022OVRL} report overall success metrics and conduct a similar robustness study with 3 random seeds. They report a similar standard deviation for OVRL of 2.7\% in overall success rate (\ourmethod + \ovrlgd is 2.4\%) and 1.7\% in SPL (\ourmethod + \ovrlgd is 0.7\%).
\begin{table}[t]
\centering
\caption{\textbf{Multi-seed runs.} Mean and standard deviation of three runs with random seeds.}
\resizebox{\textwidth}{!}{
\begin{tabular}{l c c c c c c c c}
    \toprule
    &\multicolumn{2}{c}{Overall}&\multicolumn{2}{c}{Easy}&\multicolumn{2}{c}{Medium}&\multicolumn{2}{c}{Hard}\\
    Method & Succ$\uparrow$ & SPL$\uparrow$ & Succ$\uparrow$ & SPL$\uparrow$ & Succ$\uparrow$ & SPL$\uparrow$ & Succ$\uparrow$ & SPL$\uparrow$\\
    \midrule
    \multicolumn{9}{c}{Dataset = \textit{Gibson-curved} }\\
    \midrule
    \ourmethod + \ddppogd    & 39.4 $\pm$ \small{2.5} & 24.8 $\pm$ \small{2.3} & 52.8 $\pm$ \small{2.4} & 35.2 $\pm$ \small{2.8} & 43.4 $\pm$ \small{2.6} &  27.1 $\pm$ \small{2.9} & 22.0 $\pm$ \small{2.9} & 12.0 $\pm$ \small{1.4} \\
    \ourmethod + \ovrlgd     & 56.3 $\pm$ \small{2.4} & 37.3 $\pm$ \small{0.7} & 66.6 $\pm$ \small{2.0} & 45.0 $\pm$ \small{0.7} & 61.2 $\pm$ \small{3.8} & 41.5 $\pm$ \small{1.7} & 41.1 $\pm$ \small{1.6} & 25.4 $\pm$ \small{0.8} \\
    \bottomrule
\end{tabular}
}
\label{tab:seeds-ovrl-ddppo}
\end{table}

\section{Final Distance to Goal for Top-Performing Baselines (Extending~\texorpdfstring{\secref{subsec:exp_results}}{})}
\label{subsec:dtoGoal}

\begin{table}[t]
\centering
\caption{\textbf{Distance to goal decreases with SLING.} Adding \ourmethod to previous \explore methods decreases their average distance from the final agent location to the goal.
}
\resizebox{\textwidth}{!}{
\begin{tabular}{lcccc}
\toprule
    Method & Overall Final Dist. $\downarrow$ & Easy Final Dist. $\downarrow$ & Med Final Dist. $\downarrow$ & Hard Final Dist. $\downarrow$\\
    \midrule
    DDPPO & 3.05 & 2.28 & 2.75 & 4.11\\
    SLING + \ddppogd & 2.61 & 1.68 & 2.18 & 3.96\\
    NRNS & 2.96 & 1.99 & 2.74 & 4.15\\
    SLING + \nrnsgd & 2.42 & 1.41 & 2.10 & 3.75\\
    OVRL & 2.43 & 1.58 & 2.12 & 3.59\\
    SLING + \ovrlgd & \textbf{2.17} & \textbf{1.28} & \textbf{1.70} & \textbf{3.52}\\

\bottomrule
\end{tabular}}
\vspace{-5mm}
\label{tab:dist2goal}
\end{table}

Recall, the final distance to goal metric reports the distance, from agent to goal, at the end of an episode. This metric is averaged across test episodes and reported in~\tabref{tab:dist2goal}.

Across DDPPO, NRNS, and OVRL, consistent trends hold. First, \ourmethod significantly reduces the final distance to goal. Next, the final distance to goal is much lower than the initial distance to the goal. 
As stated in averages, base OVRL starts $\sim$2.25m from the goal in easy episodes (1.5-3m) and reaches 1.58m from it, starts $\sim$4m from the goal in medium episodes (3-5m) and reaches 2.12m from it, and starts $\sim$7.5m from the goal in hard episodes (5-10m) and reached 3.59m from it.
The final trend we find is that the final distance to goal is within range of last-mile navigation. Showing that \exploit is a challenge for many previous methods.

\begin{table}[ht]
\centering
\caption{\textbf{Results testing tolerance towards.} Adding an increasing `stop' budget causes the agent to perform better. This shows that being able to recover from mistakes has great potential to improve navigation success.
$^\dagger$denotes that we edited the NRNS implementation to prevent redundant nodes from being added to the topological map. This leads to clear gains at no cost.}
\begin{tabular}{lcc}
\toprule
    `Stop' action budget & Overall Success $\uparrow$ & Overall SPL $\uparrow$\\
    \midrule
    0 from~\cite{hahn2021no} & 21.7 & 8.1 \\
    0 (reproduced$^\dagger$) & 27.8 & 10.7\\ 
    1&50.8  &17.0  \\
    2&68.4  &21.4 \\
    3&81.2  &25.1\\
    4&90.9  &28.3 \\
    5&96.2  &30.0  \\
    6&98.8  &31.0  \\
    7&99.8  &31.2 \\
    8&100.0   &31.3 \\

\bottomrule
\end{tabular}
\label{tab:stop-hammertime}
\end{table}
\section{Potential of Last-Mile Navigation -- `Stop' Budget Study (Extending \texorpdfstring{\secref{sec:related} and}{} \texorpdfstring{\secref{subsec:switches}}))}
\label{subsec:stop}

Recall that the image-goal navigation task can be completed either by calling the `stop' action, or having the agent reach the maximum number of steps in an episode. In this study, we evaluate if \exploit is a prominent error mode for image-goal navigation like it has been shown for other datasets and tasks~\cite{chattopadhyay2021robustnav,wani2020multion,ye2021auxiliary}. Following the corresponding study for multi-object navigation~\cite{wani2020multion}, we study the performance of NRNS~\cite{hahn2021no} as we increase the budget of the `stop' action errors. This stop budget allows the agent to continue last-mile navigation beyond a hard failure, until this `stop' budget is exhausted. As shown in~\tabref{tab:stop-hammertime}, with just a budget of one, success increased from $28\%$ to $51\%$. This shows that improving the last-mile of navigation and recovering from mistakes has immense potential that \ourmethod taps into.
\section{SLING with Panoramic Images (Extending~\texorpdfstring{\secref{subsec:exp_results}}{})}
\label{subsec:pan}
 Image-Goal navigation performance is highly correlated to the field-of-view (FoV) of the agent. This is intuitive as an agent that sees more about the environment and associated context in one observation will do better. Methods like NTS~\cite{chaplotNeuralTopologicalSLAM2020} and VGM~\cite{kwon2021visual} operate on panoramic observations and enjoy this advantage. However, other methods benchmarked in most prior works~\cite{yadav2022OVRL,wijmans2019dd,al2022ZER,hahn2021no} and ours operated on non-panoramic images. In order to have a fair comparison to methods that use panoramic images, we retrain OVRL with panoramic images and then apply \ourmethod to this new model. To get SLING to work with panoramic images, we take the front-facing subsection of the panoramic goal and agent image and give it to \ourmethod. For this experiment, we once again utilize the start and goal images on the Gibson-curved split, but with panoramic images. The results of this experiment are shown in~\tabref{tab:main-pan} where we demonstrate that utilizing \ourmethod with \ovrlgd yields the overall state-of-the-art on image-goal navigation while utilizing panoramic images. Notably, SLING improves the overall success and SPL of OVRL by 1.9\% and 1.0\% respectively (rows 2 and 3).

\begin{table}[ht]
\centering
\caption{\textbf{Results on the Gibson-Curved Panoramic dataset.} Adding SLING to OVRL allows us to improve their model to yield the new state-of-the-art when panoramic images are used for the image-goal navigation task.}
\resizebox{\textwidth}{!}{
\begin{tabular}{l@{\hskip 4mm} l c c c c c c c c}
    \toprule
    &&\multicolumn{2}{c}{Overall}&\multicolumn{2}{c}{Easy}&\multicolumn{2}{c}{Medium}&\multicolumn{2}{c}{Hard}\\
    & Method & Succ$\uparrow$ & SPL$\uparrow$ & Succ$\uparrow$ & SPL$\uparrow$ & Succ$\uparrow$ & SPL$\uparrow$ & Succ$\uparrow$ & SPL$\uparrow$\\
    \midrule
    1 & VGM~\cite{kwon2021visual} & 74 & 51 & 81 & 46 & 79 & 60 & 62 & \textbf{47}\\
    2 & OVRL~\cite{yadav2022OVRL} & 76.7 & 59.4 & 88.8 & 71.8 & 78.6 & 62.9 & 62.9 & 43.6\\
    3 & SLING + \ovrlgd & \textbf{78.6} & \textbf{60.4} & \textbf{90.1} & \textbf{72.9} & \textbf{82.1} & \textbf{65.0} & \textbf{63.7} & 43.4\\
    \bottomrule
\end{tabular}
}
\vspace{-3mm}
\label{tab:main-pan}
\end{table}

\section{Limitations (Extending~\texorpdfstring{\secref{sec:conclusion}}{})}
\label{subsec:supplimits}
\textit{First}, we rely on correspondences \ie mistakes in keypoint feature extraction or matching failures directly lead to errors in predicted actions (neural features~\cite{detone2018superpoint} reduce this effect). \textit{Second}, as we add structure to the \exploit problem, we also add design parameters like distance threshold $d_{\text{th}}$ and correspondence threshold $n_{\text{th}}$. The latter we tuned depending on the size of the image (640$\times$480 in NRNS and 128$\times$128 in OVRL). \textit{Third}, currently in \ourmethod, we utilize only the agent's current observation for estimating distance and heading. Using temporal smoothening could make our prediction more robust.

Following the baselines proposed in the benchmark~\cite{hahn2021no}, we also assume access to depth and pose sensors. Studies in~\cite{al2022ZER} also show improvements when using depth and pose sensors.
For physical experiments, the robot comes equipped with an inexpensive depth camera and uses SLAM~\cite{murORB2} for pose estimations.
While we demonstrate robustness to sensor noise (\tabref{tab:ablation-curved}), in future work, we could try relaxing this assumption with a depth prediction module.

Pertinent to physical experiments, other limitations are:\\
(1) Errors in pose prediction when the keypoints are located in a small area of the image. However, this can be fixed heuristically with the exploit$\rightarrow$explore switch.\\
(2) Large depth noises; this could be managed with various denoising techniques~\cite{camplani2012efficient,shen2013layer}.\\
(3) \ourmethod must directly observe the goal image in order to have enough overlap to navigate to it. Because the current image-goal navigation success criterion only requires the agent to be within 1 meter from the goal, we can not take full advantage of the task definition. However, we assert that looking at the image-goal would be more aligned with how a human would attempt image-goal navigation.\\
\end{document}